\documentclass[letterpaper]{article}
\usepackage[preprint]{poseviloc-preprint}
\usepackage[hyphens]{url}
\usepackage{graphicx}
\usepackage{natbib}
\usepackage{caption}
\usepackage{algorithm,algorithmic}
\usepackage{amsmath,amssymb}
\usepackage{booktabs,multirow,makecell,tabularx,array}
\usepackage[table]{xcolor}
\usepackage{verbatim}
\usepackage{newfloat,listings}
\usepackage[hidelinks,unicode,pdfencoding=auto]{hyperref}
\DeclareCaptionStyle{ruled}{labelfont=normalfont,labelsep=colon,strut=off}
\floatstyle{ruled}
\newfloat{listing}{tb}{lst}{}
\floatname{listing}{Listing}
\definecolor{TopOne}{RGB}{190,30,45}
\definecolor{TopTwo}{RGB}{30,90,170}
\definecolor{TopThree}{RGB}{25,135,85}

\title{PosEviLoc: Position-Conditioned Spatial Evidence for Language-Based 3D Localization}
\author{Tianyi Shang\textsuperscript{1}\equalcontrib,
  Yike Shi\textsuperscript{2}\equalcontrib,
  Zhenyu Li\textsuperscript{3}\corresponding}
\affiliations{
  \textsuperscript{1}Purdue University\\
  \textsuperscript{2}Xi'an Jiaotong-Liverpool University\\
  \textsuperscript{3}Qilu University of Technology (Shandong Academy of Sciences)\\
  \href{mailto:shangt@purdue.edu}{shangt@purdue.edu},
  \href{mailto:yike.shi22@student.xjtlu.edu.cn}{yike.shi22@student.xjtlu.edu.cn},
  \href{mailto:lizhenyu@qlu.edu.cn}{lizhenyu@qlu.edu.cn}
}
\hypersetup{
  pdftitle={PosEviLoc: Position-Conditioned Spatial Evidence for Language-Based 3D Localization},
  pdfauthor={Tianyi Shang; Yike Shi; Zhenyu Li},
  pdfsubject={Main paper and appendix},
  pdfkeywords={3D localization, point clouds, language, spatial evidence}
}
\begin{document}
\maketitle
\pagestyle{plain}
\thispagestyle{plain}
\begin{abstract}
Language-based 3D localization retrieves the point-cloud submap containing a target position from descriptions of nearby objects and their spatial relations. Existing methods typically compress queries and submaps into global descriptors, potentially obscuring object-level semantics and cross-description spatial coherence. We propose Position-Conditioned Evidence Localization (PosEviLoc), a query-position-aware framework for coarse text-to-point-cloud localization. Instead of relying on global matching, PosEviLoc evaluates each candidate submap using explicit semantic and spatial evidence. It models direction as a relation jointly determined by an object position and a hypothetical query position. The resulting Query-Position Spatial Evidence Field (QSEF) measures the fraction of query descriptions supported at each hypothetical position, explicitly capturing their agreement without using the ground-truth query pose to construct the evidence field. A Multi-Level Evidence Readout (MER) summarizes this evidence in a compact representation, which a lightweight MLP converts into a retrieval score. Across five benchmarks, PosEviLoc outperforms MNCL by an average of 17 percentage points in Recall@1. When used as a plug-and-play reranker, it improves MNCL by an average of 16 percentage points. Moreover, PosEviLoc introduces substantially fewer parameters and achieves faster inference speed than existing methods.
\end{abstract}
\begin{figure}
    \centering
    \includegraphics[width=0.9\linewidth]{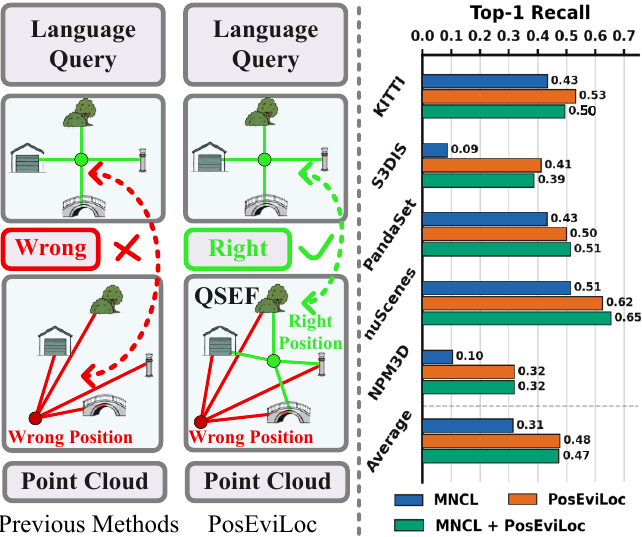}
    \caption{Left: Comparison between PosEviLoc and previous methods. Right: Top-1 recall of MNCL, PosEviLoc, and MNCL enhanced with PosEviLoc as a plug-in reranker across five datasets.}
    \label{fig:abstract}
\end{figure}
\section{Introduction}
\label{sec:introduction}
Language-based 3D localization aims to identify a target position in a
large-scale 3D environment from descriptions of nearby objects and their
spatial relations. For example, given a query such as ``the target is east of a gray building and north of a traffic sign,'' the system should retrieve the point-cloud submap containing a location that satisfies both descriptions.
This task provides an intuitive interface for applications such as autonomous
navigation, location retrieval, and embodied spatial reasoning.

Most existing methods formulate this task as cross-modal representation learning. They encode a language query and each candidate point-cloud submap into a shared embedding space and rank the candidates according to the similarity between their global descriptors~\citep{xia2024text2loc, xu2025cmmloc, liu2025text}. Representative methods, such as Text2Loc, CMMLoc, and MNCL, commonly follow a coarse-to-fine pipeline: the first stage retrieves candidate submaps from a gallery, while the second stage estimates a more precise position within the retrieved submaps. This paradigm enables efficient retrieval from large-scale submap galleries and has achieved substantial progress. However, compressing an entire query and submap into a single descriptor may obscure the fine-grained evidence required for language-based localization. A high global similarity does not necessarily indicate that the candidate contains objects with the requested classes and colors, nor does it ensure that multiple descriptions admit a spatially coherent explanation. 

This limitation is particularly important for directional language. Relations such as \textit{east}, \textit{west}, \textit{north}, and \textit{south} are not intrinsic attributes of scene objects. Instead, they depend on the relative configuration of an object and the query position. The same object may be east of one location but west of another. Assigning a static direction label to each object therefore removes the position-dependent nature of spatial language. Moreover, independently matching individual descriptions is insufficient: multiple class--color--direction descriptions may each be supported somewhere in a candidate submap while corresponding to mutually inconsistent query positions. A reliable retrieval model should therefore determine whether the objects in a candidate submap jointly support as many query descriptions as possible at a shared location.

To address these limitations, we propose Position-Conditioned Evidence Localization (PosEviLoc), a spatial evidence retrieval framework for coarse text-to-point-cloud localization. Unlike conventional retrieval methods, PosEviLoc does not produce a global descriptor for query--submap matching. Instead, it evaluates each candidate submap through explicit object-level semantic and spatial evidence. PosEviLoc improves only the first-stage submap retrieval, while the original fine-localization module is retained without modification and is not considered part of our contribution.

The core of PosEviLoc is a Query-Position Spatial Evidence Field (QSEF), which
defines a two-dimensional grid of hypothetical query positions for each
candidate submap, as shown in Figure~\ref{fig:abstract} left. At each grid position, object directions are dynamically
determined from their relative offsets to the hypothetical query position.
QSEF then measures the fraction of query descriptions supported by objects
with the required class, color, and position-conditioned direction,
explicitly capturing whole-query spatial coherence without using the
ground-truth pose.

A Multi-Level Evidence Readout (MER) further summarizes complementary semantic
and spatial evidence, including class--color compatibility, shared-position
soft agreement, exact spatial feasibility, and the spatial extent of valid
support. These statistics form a compact evidence representation that is
mapped to a retrieval score by a lightweight multilayer perceptron.

PosEviLoc can operate independently or serve as a plug-and-play reranking
module for existing methods without modifying or retraining their encoders. As shown in Figure~\ref{fig:abstract} (right), PosEviLoc surpasses MNCL by an average of 17 Recall@1 points across five benchmarks and improves it by 16 points as a reranker.

Our main contributions are summarized as follows:
\begin{itemize}
\item We propose PosEviLoc, which formulates coarse text-to-point-cloud
localization as position-conditioned evidence evaluation, explicitly
verifying whether candidate objects jointly support the query.

\item We introduce QSEF and MER. QSEF evaluates class, color, and direction
evidence over hypothetical query positions, while MER summarizes semantic
compatibility, soft agreement, and spatial feasibility for retrieval.

\item PosEviLoc achieves the best Recall@1 on all five benchmarks and, as a
plug-and-play reranker for six methods, improves 89 of 90 reported
R@1--R@3 results. PosEviLoc also achieves faster inference with lower computational overhead.
\end{itemize}

\begin{figure*}
    \centering
    \includegraphics[width=0.9\linewidth]{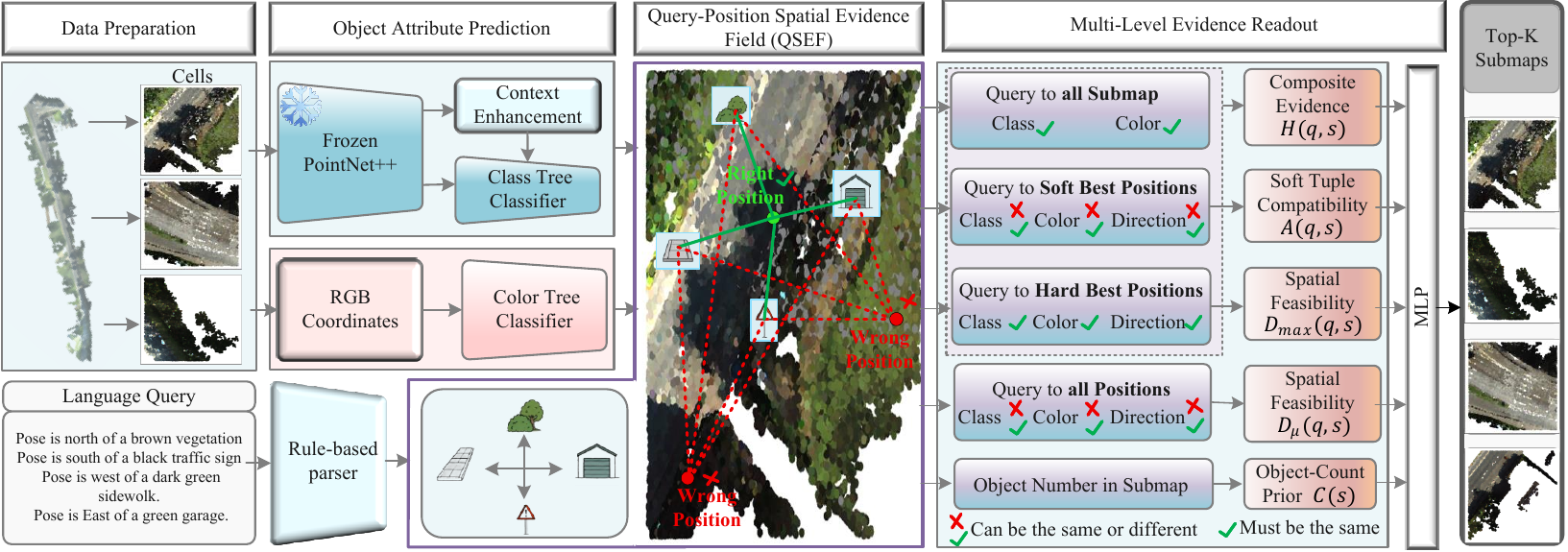}
    \caption{Overall architecture of the proposed PosEviLoc framework.}
    \label{figure:method}
\end{figure*}
\section{Related Work}

\subsection{LiDAR Place Recognition}

LiDAR place recognition (LPR) retrieves previously observed locations from
3D scans and is robust to illumination and appearance changes. Early methods
used handcrafted geometric descriptors. M2DP derives a compact signature from
multi-view point-cloud projections~\cite{he2016m2dp}, while Scan Context
employs a polar grid for efficient yaw-aware retrieval~\cite{kim2018scan}.
However, these descriptors remain sensitive to sparsity, occlusion, and
structural repetition.

Learning-based methods instead learn global descriptors from data.
PointNetVLAD combines PointNet with NetVLAD~\cite{uy2018pointnetvlad}, while
LPD-Net enhances local geometry through adaptive feature extraction and graph
aggregation~\cite{liu2019lpd}. MinkLoc3D uses sparse voxel
convolutions~\cite{komorowski2021minkloc3d}, whereas SOE-Net and PPT-Net adopt
orientation-aware attention and multi-scale Transformer
aggregation~\cite{xia2021soe,hui2021pyramid}. LoGG3D-Net further improves
descriptor repeatability through local feature consistency~\cite{vidanapathirana2022logg3d}.

Recent methods explore hybrid representations and stronger robustness. CASSPR
fuses point-wise and sparse-voxel features through cross-attention~\cite{xia2023casspr},
while BEVPlace performs rotation-equivariant retrieval
from bird's-eye-view projections~\cite{luo2023bevplace}. ForestLPR and
HOTFormerLoc extend LPR to forests and ground--aerial settings~\cite{shen2025forestlpr,griffiths2025hotformerloc},
whereas C-LaV improves robustness under adverse weather through latent
denoising~\cite{cao2026c}. Overall, LPR has evolved from handcrafted signatures toward
learned multi-scale and condition-robust geometric representations.

\subsection{Cross-Modal Place Recognition}
\label{sec:cross_modal_place_recognition}

Natural-language-based place recognition enables users to specify target
locations through nearby objects, visual attributes, and spatial relations.
Text2Pos introduced text-to-point-cloud localization and the KITTI360Pose
benchmark, establishing the coarse-to-fine paradigm of retrieving candidate
submaps before estimating the target position within the selected
submap~\cite{kolmet2022text2pos}. Most subsequent methods follow this design.
RET models relations between textual hints and 3D
instances~\cite{wang2023text}, while Text2Loc combines hierarchical text
aggregation, contrastive text--submap learning, and matching-free fine
localization~\cite{xia2024text2loc}.

MambaPlace uses modality-specific and cross-modal Mamba modules to capture
long-range dependencies~\cite{shang2024mambaplace}. Des4Pos models multi-scale spatial relations with a bidirectional LSTM and reduces the modality gap using CLIP priors~\cite{shang2026vehicle}.
FourierPlace introduces frequency-domain
point-cloud representations, Fourier-gated attention, and hierarchical
language modeling~\cite{shang2025fourierplace}. PMSH \cite{feng2025partially} handles partially matched submaps by propagating uncertainty into the coarse retrieval metric. MNCL adopts multi-level
negative contrastive learning~\cite{liu2025text}, whereas CMMLoc introduces
uncertainty-aware text--scene association and directional
modeling~\cite{xu2025cmmloc}. Unlike full coarse-to-fine pipelines, VLM-Loc
focuses on fine localization within retrieved submaps using a vision-language
model~\cite{kang2026vlm}.

\section{Method}
\label{sec:method}

\subsection{Overview}
PosEviLoc retrieves submaps by testing whether their objects jointly explain a
language query from a plausible position, rather than by matching global
descriptors (Figure~\ref{figure:method}).

A fixed rule-based parser converts the query into object descriptions
\begin{equation}
    q=\{d_i\}_{i=1}^{m}, \qquad d_i=(c_i,a_i,r_i),
    \label{eq:query_definition}
\end{equation}
where $c_i\in\mathcal{C}$, $a_i\in\mathcal{A}$, and $r_i\in\mathcal{R}$
denote class, color, and spatial relation, respectively, and
$\mathcal{R}=\{\text{north},\text{south},\text{east},\text{west},\text{on-top}\}$.
The parser is fixed across data splits and has no learnable parameters; exact
duplicate tuples are merged, leaving $m$ distinct descriptions. Each candidate
$s\in\mathcal{S}$ contains an object set $\mathcal{O}_s$.

For each candidate, PosEviLoc predicts object attributes, constructs a
Query-Position Spatial Evidence Field (QSEF), and uses a
Multi-Level Evidence Readout (MER) to map semantic and spatial evidence
to a retrieval score $z_{\theta}(q,s)$. The ground-truth query pose is used only
to identify training positives and compute evaluation metrics, never to derive
object attributes or spatial relations.

\subsection{Object Attribute Prediction}
\label{sec:object_prediction}

For each $o\in\mathcal{O}_s$, PosEviLoc predicts its class $\hat c_o$, color
$\hat a_o$, and horizontal location $\mathbf{p}_o\in\mathbb{R}^{2}$.

\paragraph{Context-aware class prediction.}
A base PointNet++ classifier provides a geometric feature
$\mathbf{x}^{\mathrm{geo}}_o$ and class posterior
$\mathbf{b}_o\in[0,1]^{|\mathcal{C}|}$. We augment them with
$\boldsymbol{\ell}_o=E_{\mathrm{layout}}(o,\mathcal{O}_s)$, which summarizes
relative position, scale, and local density, yielding
\begin{equation}
    \mathbf{h}^{\mathrm{cls}}_o=
    [\mathbf{x}^{\mathrm{geo}}_o;\mathbf{b}_o;\boldsymbol{\ell}_o].
\end{equation}
A tree ensemble then combines object and contextual evidence:
\begin{equation}
\begin{aligned}
    \mathbf{p}^{\mathrm{cls}}(o)
    &=
    \frac{1}{K_c}
    \sum_{k=1}^{K_c}
    T_k^{\mathrm{cls}}(\mathbf{h}^{\mathrm{cls}}_o),\\
    \hat c_o
    &=
    \arg\max_{c\in\mathcal{C}}
    p^{\mathrm{cls}}_c(o),
\end{aligned}
\label{eq:class_context}
\end{equation}
where $K_c$ is the number of trees and $T_k^{\mathrm{cls}}$ outputs a class
posterior.

\paragraph{RGB-based color prediction.}
Let $\mathbf{r}_o$ denote the RGB observations of object $o$. We directly feed
these RGB values to a lightweight tree-based classifier:
\begin{equation}
    \mathbf{p}^{\mathrm{col}}(o)
    =
    T^{\mathrm{col}}(\mathbf{r}_o),
    \qquad
    \hat a_o
    =
    \arg\max_{a\in\mathcal{A}}
    p^{\mathrm{col}}_a(o).
    \label{eq:color_prediction}
\end{equation}
This branch provides explicit color evidence without introducing a separate
high-capacity appearance encoder.

\subsection{Query-Position Spatial Evidence Field}
\label{sec:qsef}

The core of PosEviLoc is the {Query-Position Spatial Evidence Field
(QSEF)}, which evaluates whether the objects in a candidate submap can support
the query from a common hypothetical location. A direction is therefore
treated as a relation induced jointly by a query position and an object
position, rather than as a fixed object attribute.

We place a normalized 2D grid $G$ on each submap, where each
$\mathbf{g}\in G$ represents a hypothetical query position rather than an
observed point or object center. For object $o$, define
\begin{equation}
    \boldsymbol{\Delta}(\mathbf{g},o)
    =
    \mathbf{g}-\mathbf{p}_o
    =
    (\Delta_x,\Delta_y).
    \label{eq:relative_offset}
\end{equation}
Let $\rho(\mathbf{g},o)=\|\boldsymbol{\Delta}(\mathbf{g},o)\|_2$. The
position-conditioned direction operator is
\begin{equation}
D(\mathbf{g},o)=
\begin{cases}
\text{on-top},
& \rho(\mathbf{g},o)\leq\tau,\\
\text{east},
& \rho(\mathbf{g},o)>\tau,\ |\Delta_x|>|\Delta_y|,\ \Delta_x>0,\\
\text{west},
& \rho(\mathbf{g},o)>\tau,\ |\Delta_x|>|\Delta_y|,\ \Delta_x\leq 0,\\
\text{north},
& \rho(\mathbf{g},o)>\tau,\ |\Delta_x|\leq|\Delta_y|,\ \Delta_y>0,\\
\text{south},
& \rho(\mathbf{g},o)>\tau,\ |\Delta_x|\leq|\Delta_y|,\ \Delta_y\leq 0.
\end{cases}
\label{eq:direction_operator}
\end{equation}
Here, $\tau=0.06$ is the on-top radius in normalized submap coordinates.
We denote the resulting relation by
\begin{equation}
    \hat r_o(\mathbf{g})
    \triangleq
    D(\mathbf{g},o).
    \label{eq:position_conditioned_relation}
\end{equation}
Thus, $\hat r_o(\mathbf{g})$ is computed from $\mathbf{g}$ and
$\mathbf{p}_o$, rather than from a static object annotation or a relation label
derived from the ground-truth query pose.

For description $d_i=(c_i,a_i,r_i)$, exact support at $\mathbf{g}$ is
\begin{equation}
\begin{aligned}
    M_i^{(s)}(\mathbf{g})
    =
    \mathbf{1}\!\left[
    \exists o\in\mathcal{O}_s:\,
    \hat c_o=c_i,\,
    \hat a_o=a_i,\,
    \hat r_o(\mathbf{g})=r_i
    \right].
\end{aligned}
\label{eq:description_evidence}
\end{equation}
Here, $\mathbf{1}[\cdot]$ equals one when the enclosed condition holds and zero
otherwise. Hence, $M_i^{(s)}(\mathbf{g})=1$ only when one scene object jointly
supports the class, color, and direction required by description $i$ at
position $\mathbf{g}$. Aggregating all descriptions gives the
query-conditioned evidence field
\begin{equation}
    U_{q,s}(\mathbf{g})
    =
    \frac{1}{m}
    \sum_{i=1}^{m}
    M_i^{(s)}(\mathbf{g}).
    \label{eq:query_evidence_field}
\end{equation}
A high value indicates that many descriptions agree at the same hypothetical
query position, so QSEF captures a joint spatial explanation rather than a
count of individually compatible objects. Descriptions are evaluated
independently, allowing one physical instance to support multiple
co-referential descriptions. At a fixed $\mathbf{g}$, however, each object has
only one direction and cannot support mutually incompatible directional
requirements.

\subsection{Multi-Level Evidence Readout}
\label{sec:mer}

Given the QSEF, the Multi-Level Evidence Readout (MER) summarizes each
query--submap pair from complementary semantic and spatial perspectives. It
measures semantic compatibility, soft tuple compatibility at a shared
position, exact spatial feasibility, and the spatial extent of valid support.
In addition, we introduce a lightweight object-count prior to characterize the
overall amount of scene evidence available in each candidate submap. These
statistics preserve the interpretable structure of the QSEF while forming a
compact representation for retrieval scoring.

\paragraph{Semantic Compatibility.}
We first measure whether each description can find a semantically compatible
object without enforcing directional consistency:
\begin{equation}
\begin{aligned}
I(q,s)
&=
\frac{1}{m}
\sum_{i=1}^{m}
\max_{o\in\mathcal{O}_s}
\frac{1}{2}
\left(
\mathbf{1}[\hat c_o=c_i]
+
\mathbf{1}[\hat a_o=a_i]
\right).
\end{aligned}
\label{eq:identity_compatibility}
\end{equation}
Because both indicators are evaluated on the same object, a full score requires
the class and color to match on the same instance, whereas a match on either
attribute receives partial credit. Thus, $I(q,s)$ provides a tolerant semantic
signal when one predicted attribute is unreliable, while avoiding the
combination of unrelated class and color matches from different objects.

\paragraph{Soft Tuple Compatibility.}
We next incorporate the position-conditioned direction while retaining partial
credit for incomplete tuple matches. For description $d_i$, object $o$, and
hypothetical query position $\mathbf{g}$, we define
\begin{equation}
\begin{aligned}
\alpha_i(\mathbf{g},o)
=
\frac{1}{3}
\bigl(
&\mathbf{1}[\hat c_o=c_i]
+
\mathbf{1}[\hat a_o=a_i]\\
&+
\mathbf{1}[\hat r_o(\mathbf{g})=r_i]
\bigr).
\end{aligned}
\label{eq:tuple_compatibility}
\end{equation}
The position-wise and submap-level compatibility scores are defined as
\begin{equation}
\widetilde A(q,s;\mathbf{g})
=
\frac{1}{m}
\sum_{i=1}^{m}
\max_{o\in\mathcal{O}_s}
\alpha_i(\mathbf{g},o),
\label{eq:positionwise_attribute_compatibility}
\end{equation}
and
\begin{equation}
A(q,s)
=
\max_{\mathbf{g}\in G}
\widetilde A(q,s;\mathbf{g}).
\label{eq:attribute_compatibility}
\end{equation}
All descriptions are evaluated at the same shared position $\mathbf{g}$,
preventing their directional requirements from being satisfied at unrelated
description-specific positions. Since $\alpha_i(\mathbf{g},o)$ assigns partial
credit to incomplete class--color--direction tuples, $A(q,s)$ measures graded
compatibility rather than requiring all three attributes to match exactly. It
therefore preserves useful partial evidence when one predicted attribute is
incorrect.
\newcommand{\topone}[1]{\textbf{#1}}
\newcommand{\toptwo}[1]{\underline{#1}}
\newcommand{\topthree}[1]{\textit{#1}}

\begin{table*}[t]
    \centering

    {%
    \small
    \setlength{\tabcolsep}{1mm}

    \begin{tabular}{
        @{}
        c
        | c c c
        | c c c
        | c c c
        | c c c
        | c c c
        @{}
    }
        \toprule

        \multicolumn{1}{c|}{%
            \multirow{2}{*}{\textbf{Method}}%
        }
        & \multicolumn{3}{c|}{\textbf{KITTI360Pose}}
        & \multicolumn{3}{c|}{\textbf{S3DISPose}}
        & \multicolumn{3}{c|}{\textbf{PandaSetPose}}
        & \multicolumn{3}{c|}{\textbf{nuScenesPose}}
        & \multicolumn{3}{c}{\textbf{NPM3DCoursePose}}
        \\

        \cmidrule(lr){2-4}
        \cmidrule(lr){5-7}
        \cmidrule(lr){8-10}
        \cmidrule(lr){11-13}
        \cmidrule(lr){14-16}

        & \textbf{R@1} & \textbf{R@2} & \textbf{R@3}
        & \textbf{R@1} & \textbf{R@2} & \textbf{R@3}
        & \textbf{R@1} & \textbf{R@2} & \textbf{R@3}
        & \textbf{R@1} & \textbf{R@2} & \textbf{R@3}
        & \textbf{R@1} & \textbf{R@2} & \textbf{R@3}
        \\

        \midrule

        Text2Loc
        & 0.30 & 0.43 & 0.51
        & \topthree{0.09} & 0.17 & 0.25
        & \toptwo{0.46} & \topone{0.64} & \topone{0.72}
        & 0.55 & 0.67 & 0.72
        & \topthree{0.18} & \topthree{0.30} & 0.37
        \\

        FourierPlace
        & \topthree{0.33} & \topthree{0.48} & \topthree{0.55}
        & 0.05 & 0.10 & 0.13
        & 0.29 & 0.38 & 0.46
        & \toptwo{0.61} & \topone{0.76} & \topone{0.80}
        & 0.10 & 0.22 & 0.28
        \\

        Des4Pos
        & 0.31 & 0.44 & 0.52
        & 0.08 & \topthree{0.18} & \toptwo{0.27}
        & 0.39 & 0.53 & 0.60
        & 0.54 & 0.67 & 0.72
        & \toptwo{0.19} & \toptwo{0.41} & \toptwo{0.49}
        \\

        MambaPlace
        & 0.30 & 0.44 & 0.51
        & \toptwo{0.10} & \toptwo{0.19} & \topthree{0.26}
        & 0.19 & 0.29 & 0.36
        & 0.53 & 0.62 & 0.71
        & 0.14 & 0.26 & \topthree{0.39}
        \\

        MNCL
        & \toptwo{0.43} & \toptwo{0.58} & \toptwo{0.66}
        & \topthree{0.09} & 0.15 & 0.22
        & \topthree{0.43} & \topthree{0.56} & \topthree{0.64}
        & 0.51 & 0.63 & 0.70
        & 0.10 & 0.16 & 0.19
        \\

        CMMLoc
        & 0.32 & 0.46 & 0.54
        & \toptwo{0.10} & \toptwo{0.19} & \topthree{0.26}
        & 0.27 & 0.43 & 0.54
        & \topthree{0.58} & \topthree{0.71} & \topthree{0.75}
        & 0.11 & 0.24 & 0.34
        \\

        \midrule

        \textbf{PosEviLoc (Ours)}
        & \topone{0.53} & \topone{0.65} & \topone{0.71}
        & \topone{0.41} & \topone{0.53} & \topone{0.60}
        & \topone{0.50} & \toptwo{0.62} & \toptwo{0.67}
        & \topone{0.62} & \toptwo{0.72} & \toptwo{0.77}
        & \topone{0.32} & \topone{0.46} & \topone{0.56}
        \\

        \bottomrule
    \end{tabular}
    }
    \caption{
        Coarse-retrieval performance on five language-based 3D localization
        benchmarks, measured by Recall@$K$. The best, second-best, and
        third-best results are indicated by boldface, underlining, and
        italics, respectively.
    }
    \label{tab:main_results}
\end{table*}

\paragraph{Spatial Feasibility.}
To complement the partial evidence captured by the graded compatibility score,
we compute
\begin{equation}
D_{\max}(q,s)
=
\max_{\mathbf{g}\in G}
U_{q,s}(\mathbf{g}),
\label{eq:peak_feasibility}
\end{equation}
which represents the largest fraction of descriptions that can be exactly
satisfied at a single shared position. Unlike $A(q,s)$, which permits partial
tuple matches, $D_{\max}(q,s)$ counts only complete
class--color--direction matches and therefore directly measures whole-query
spatial coherence.

We further measure the spatial extent of exact support as
\begin{equation}
D_{\mu}(q,s)
=
\frac{1}{|G|}
\sum_{\mathbf{g}\in G}
U_{q,s}(\mathbf{g}).
\label{eq:average_feasibility}
\end{equation}
Here, $D_{\max}$ captures the strongest shared explanation at a single
hypothetical position, whereas $D_{\mu}$ complements this peak value by
describing how broadly exact support is distributed across the latent query
position space. Together, they distinguish sharp and localized agreement from
diffuse support spread over a larger region.

\paragraph{Object-Count Prior.}
We additionally encode the number of detected objects in each candidate submap
as a coarse scene-density prior:
\begin{equation}
C(s)
=
\frac{
\log\left(1+\left|\mathcal{O}_s\right|\right)
}{
\log\left(1+N_{\max}\right)
},
\label{eq:object_count_prior}
\end{equation}
where $N_{\max}$ is the maximum number of detected objects in any training
submap. The logarithm reduces the influence of unusually dense submaps, while
the denominator normalizes the feature to a scale comparable to those of the
remaining evidence measures. Thus, $C(s)$ provides a query-independent
estimate of the amount of available scene evidence without allowing the object
count to dominate the retrieval representation.

\paragraph{Composite Evidence Summary.}
The preceding statistics capture distinct but complementary properties of a
candidate submap. We combine the three principal compatibility measures as
\begin{equation}
H(q,s)
=
I(q,s)
+
A(q,s)
+
D_{\max}(q,s).
\label{eq:hybrid_summary}
\end{equation}
This equal-weight summary provides the shallow scorer with an explicit
monotonic combination of semantic compatibility, soft shared-position support,
and exact spatial feasibility. We do not treat $H(q,s)$ as an independent
source of information; rather, it is a linear reparameterization of the main
evidence coordinates.

\paragraph{Retrieval Representation.}
Because $I(q,s)=H(q,s)-A(q,s)-D_{\max}(q,s)$, semantic compatibility remains
recoverable from the retained components. We use the representation
\begin{equation}
\boldsymbol{\phi}(q,s)
=
\bigl[
H(q,s),
A(q,s),
D_{\max}(q,s),
D_{\mu}(q,s),
C(s)
\bigr].
\label{eq:retrieval_representation}
\end{equation}
The retrieval score is then computed as
\begin{equation}
z_{\theta}(q,s)
=
f_{\theta}\!\left(
\boldsymbol{\phi}(q,s)
\right),
\label{eq:retrieval_score}
\end{equation}
where $f_{\theta}$ is a two-hidden-layer MLP with parameters $\theta$, ReLU
activations, and dropout. We train it with listwise cross-entropy to rank the
positive submap above hard and randomly sampled negatives. The
scorer calibrates the relative contributions of semantic, directional, and
spatial evidence across different query patterns.

After scoring all candidate submaps, PosEviLoc ranks them in descending order
of $z_{\theta}(q,s)$ and returns the top-$K$ results:
\begin{equation}
\widehat{\mathcal{S}}_{K}(q)
=
\operatorname*{TopK}_{s\in\mathcal{S}}
z_{\theta}(q,s).
\label{eq:topk_retrieval}
\end{equation}
For plug-and-play use, PosEviLoc reranks only the Top-25 submaps returned by
the base retriever.

\section{Experiments}
\subsection{Dataset Preparation}
We evaluate PosEviLoc on five language-based 3D localization datasets covering outdoor driving, indoor rooms, and urban mobile-mapping environments. Each query describes a sampled pose using nearby objects, semantic attributes, and relative spatial relations, while the task is evaluated as coarse cell retrieval.

\paragraph{KITTI360Pose.}
KITTI360Pose is built from nine KITTI-360 outdoor driving sequences. Following the official setting, we use 30-m cells with 10-m spacing and the original split. Its queries mainly involve static roadside structures and strong cardinal-direction cues.

\paragraph{PandaSetPose.}
PandaSetPose covers 100 PandaSet driving sequences using 30-m cells with 10-m spacing. Compared with KITTI360Pose, it contains more dynamic traffic participants, including vehicles and pedestrians.

\paragraph{S3DISPose.}
S3DISPose contains 272 aligned indoor rooms. Unlike the outdoor datasets, each room is treated as an adaptive-size cell, and queries describe indoor elements such as walls, furniture, and clutter.

\paragraph{nuScenesPose.}
nuScenesPose is constructed from 850 urban driving scenes with 30-m cells and 10-m spacing. It provides diverse urban traffic layouts for evaluating localization in dense urban environments.

\paragraph{NPM3DCoursePose.}
NPM3DCoursePose covers mobile laser scans from Lille and Paris and comprises 91 cells of 20 m with 10-m spacing. Unlike the driving datasets, it represents large-scale European street environments acquired through mobile mapping.

\newcommand{\gainresult}[2]{%
    \shortstack[c]{%
        #1\\[-0.45ex]
        \textbf{(#2)}%
    }%
}

\begin{table*}[t]
    \centering

    \begingroup
    \fontsize{9pt}{10.5pt}\selectfont
    \normalfont

    \setlength{\tabcolsep}{0.7pt}
    \renewcommand{\arraystretch}{1.08}

    \begin{tabular*}{\textwidth}{
        @{\extracolsep{\fill}}
        >{\centering\arraybackslash}p{1.55cm} |
        c c c |
        c c c |
        c c c |
        c c c |
        c c c
        @{}
    }
        \toprule

        \multicolumn{1}{c|}{%
            \multirow{2}{*}{\textbf{Method}}%
        }
        &
        \multicolumn{3}{c|}{%
            \shortstack[c]{\textbf{KITTI360Pose}}%
        }
        &
        \multicolumn{3}{c|}{%
            \shortstack[c]{\textbf{S3DISPose}}%
        }
        &
        \multicolumn{3}{c|}{%
            \shortstack[c]{\textbf{PandaSetPose}}%
        }
        &
        \multicolumn{3}{c|}{%
            \shortstack[c]{\textbf{nuScenesPose}}%
        }
        &
        \multicolumn{3}{c}{%
            \shortstack[c]{\textbf{NPM3DCoursePose}}%
        }
        \\

        \cmidrule(lr){2-4}
        \cmidrule(lr){5-7}
        \cmidrule(lr){8-10}
        \cmidrule(lr){11-13}
        \cmidrule(lr){14-16}

        &
        \textbf{R@1} & \textbf{R@2} & \textbf{R@3}
        &
        \textbf{R@1} & \textbf{R@2} & \textbf{R@3}
        &
        \textbf{R@1} & \textbf{R@2} & \textbf{R@3}
        &
        \textbf{R@1} & \textbf{R@2} & \textbf{R@3}
        &
        \textbf{R@1} & \textbf{R@2} & \textbf{R@3}
        \\

        \midrule

        \textbf{Text2Loc$^{+}$}
        & \gainresult{0.45}{+0.15}
        & \gainresult{0.59}{+0.16}
        & \gainresult{0.66}{+0.15}
        & \gainresult{0.42}{+0.33}
        & \gainresult{0.54}{+0.37}
        & \gainresult{0.61}{+0.36}
        & \gainresult{0.54}{+0.08}
        & \gainresult{0.66}{+0.02}
        & \gainresult{0.73}{+0.01}
        & \gainresult{0.65}{+0.10}
        & \gainresult{0.76}{+0.09}
        & \gainresult{0.80}{+0.08}
        & \gainresult{0.32}{+0.14}
        & \gainresult{0.46}{+0.16}
        & \gainresult{0.56}{+0.19}
        \\
        \addlinespace[1.2pt]

        \shortstack[c]{%
            \textbf{FourierP$^{+}$}%
        }
        & \gainresult{0.46}{+0.13}
        & \gainresult{0.61}{+0.13}
        & \gainresult{0.68}{+0.13}
        & \gainresult{0.35}{+0.30}
        & \gainresult{0.44}{+0.34}
        & \gainresult{0.49}{+0.36}
        & \gainresult{0.52}{+0.23}
        & \gainresult{0.62}{+0.24}
        & \gainresult{0.68}{+0.22}
        & \gainresult{0.65}{+0.04}
        & \gainresult{0.76}{+0.00}
        & \gainresult{0.82}{+0.02}
        & \gainresult{0.32}{+0.22}
        & \gainresult{0.46}{+0.24}
        & \gainresult{0.56}{+0.28}
        \\
        \addlinespace[1.2pt]

        \textbf{Des4Pos$^{+}$}
        & \gainresult{0.46}{+0.15}
        & \gainresult{0.60}{+0.16}
        & \gainresult{0.67}{+0.15}
        & \gainresult{0.42}{+0.34}
        & \gainresult{0.54}{+0.36}
        & \gainresult{0.61}{+0.34}
        & \gainresult{0.54}{+0.15}
        & \gainresult{0.66}{+0.13}
        & \gainresult{0.72}{+0.12}
        & \gainresult{0.66}{+0.12}
        & \gainresult{0.75}{+0.08}
        & \gainresult{0.81}{+0.09}
        & \gainresult{0.32}{+0.13}
        & \gainresult{0.46}{+0.05}
        & \gainresult{0.56}{+0.07}
        \\
        \addlinespace[1.2pt]

        \shortstack[c]{%
            \textbf{MambaP$^{+}$}%
        }
        & \gainresult{0.45}{+0.15}
        & \gainresult{0.59}{+0.15}
        & \gainresult{0.66}{+0.15}
        & \gainresult{0.41}{+0.31}
        & \gainresult{0.52}{+0.33}
        & \gainresult{0.58}{+0.32}
        & \gainresult{0.47}{+0.28}
        & \gainresult{0.57}{+0.28}
        & \gainresult{0.67}{+0.31}
        & \gainresult{0.66}{+0.13}
        & \gainresult{0.77}{+0.15}
        & \gainresult{0.82}{+0.11}
        & \gainresult{0.34}{+0.20}
        & \gainresult{0.47}{+0.21}
        & \gainresult{0.56}{+0.17}
        \\
        \addlinespace[1.2pt]

        \textbf{MNCL$^{+}$}
        & \gainresult{0.50}{+0.07}
        & \gainresult{0.64}{+0.06}
        & \gainresult{0.70}{+0.04}
        & \gainresult{0.39}{+0.30}
        & \gainresult{0.49}{+0.34}
        & \gainresult{0.58}{+0.36}
        & \gainresult{0.51}{+0.08}
        & \gainresult{0.63}{+0.07}
        & \gainresult{0.69}{+0.05}
        & \gainresult{0.65}{+0.14}
        & \gainresult{0.76}{+0.13}
        & \gainresult{0.82}{+0.12}
        & \gainresult{0.32}{+0.22}
        & \gainresult{0.46}{+0.30}
        & \gainresult{0.56}{+0.37}
        \\
        \addlinespace[1.2pt]

        \textbf{CMMLoc$^{+}$}
        & \gainresult{0.46}{+0.14}
        & \gainresult{0.61}{+0.15}
        & \gainresult{0.68}{+0.14}
        & \gainresult{0.41}{+0.31}
        & \gainresult{0.49}{+0.30}
        & \gainresult{0.58}{+0.32}
        & \gainresult{0.54}{+0.27}
        & \gainresult{0.66}{+0.23}
        & \gainresult{0.72}{+0.18}
        & \gainresult{0.67}{+0.09}
        & \gainresult{0.76}{+0.05}
        & \gainresult{0.81}{+0.06}
        & \gainresult{0.32}{+0.21}
        & \gainresult{0.46}{+0.22}
        & \gainresult{0.56}{+0.22}
        \\

        \bottomrule
    \end{tabular*}

    \endgroup

    \caption{
        PosEviLoc as a top-25 plug-in reranker. The superscript “+” denotes integration with PosEviLoc. Each entry reports the reranked recall
        on the first line and the absolute gain over the corresponding
        original result in Table~\ref{tab:main_results} on the second line.}
    \label{tab:plugin_reranking}
\end{table*}

\begin{figure}
    \centering
    \includegraphics[width=1\linewidth]{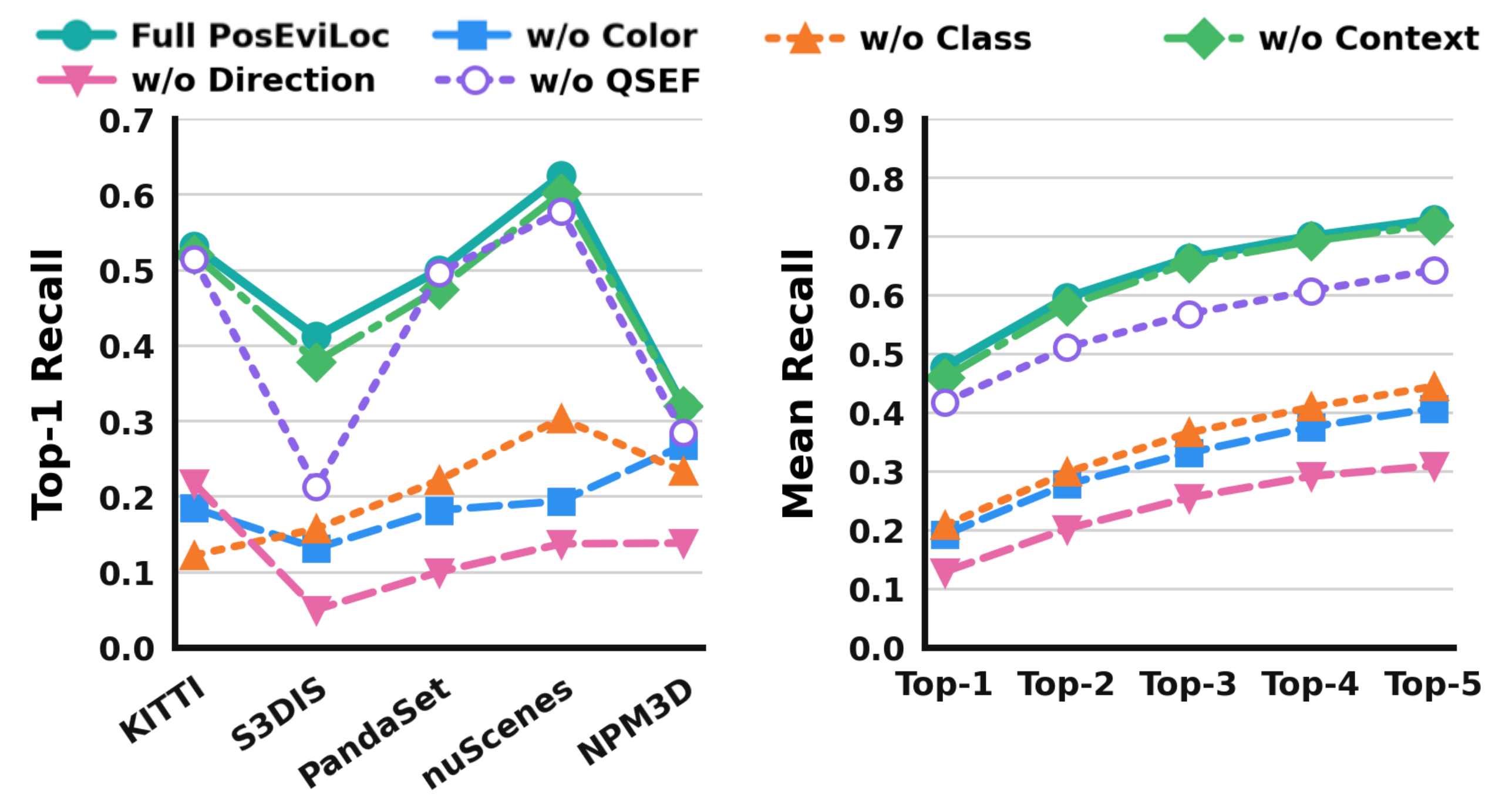}
     \caption{Component ablation. ``w/o Class/Color/Direction'' removes the corresponding symbolic cue, ``w/o Context'' disables context-aware class refinement, and ``w/o QSEF'' assigns directions at the cell center.}
    \label{fig:component_ablation}
\end{figure}


\newcommand{\mernum}[1]{#1}
\newcommand{\merbest}[1]{\textbf{#1}}

\newcommand{\lossresult}[2]{%
    \makecell[c]{%
        #1\\
        \textit{(#2)}%
    }%
}

\begin{table}[t]
    \centering

    \begingroup
    \small
    \setlength{\tabcolsep}{1.5pt}

    \begin{tabularx}{\columnwidth}{
        @{}
        >{\centering\arraybackslash}X |
        c c c c c
        @{}
    }
        \toprule

        \multicolumn{1}{c|}{\textbf{MER Variant}}
        & \textbf{KITTI}
        & \textbf{S3DIS}
        & \textbf{Panda}
        & \textbf{nuScenes}
        & \textbf{NPM3D}
        \\

        \midrule

        \textbf{PosEviLoc (Full)}
        & \merbest{0.53}
        & \merbest{0.41}
        & \merbest{0.50}
        & \merbest{0.62}
        & \merbest{0.32}
        \\

        w/o $H(q,s)$
        & \mernum{0.35}
        & \mernum{0.25}
        & \mernum{0.46}
        & \mernum{0.39}
        & \mernum{0.20}
        \\

        w/o $A(q,s)$
        & \mernum{0.36}
        & \merbest{0.41}
        & \merbest{0.50}
        & \merbest{0.62}
        & \mernum{0.29}
        \\

        w/o $D_{\max}(q,s)$
        & \mernum{0.52}
        & \mernum{0.31}
        & \mernum{0.47}
        & \mernum{0.60}
        & \mernum{0.31}
        \\

        w/o $D_{\mu}(q,s)$
        & \merbest{0.53}
        & \mernum{0.36}
        & \mernum{0.45}
        & \mernum{0.57}
        & \merbest{0.32}
        \\

        w/o $C(s)$
        & \mernum{0.50}
        & \mernum{0.31}
        & \mernum{0.41}
        & \mernum{0.40}
        & \mernum{0.29}
        \\

        \bottomrule
    \end{tabularx}

    \endgroup

    \caption{
        Leave-one-feature-out ablation of the five-dimensional MER
        representation. Recall@1 is reported, and the best result in each
        column is shown in bold.
    }
    \label{tab:mer_feature_ablation}
\end{table}

\subsection{Comparison with State of the Art}
Table~\ref{tab:main_results} shows that PosEviLoc obtains the best R@1 on all
five benchmarks. Relative to the strongest baseline on each dataset, the
absolute R@1 gains are 0.10 on KITTI360Pose, 0.31 on S3DISPose, 0.04 on
PandaSetPose, 0.01 on nuScenesPose, and 0.13 on NPM3DCoursePose. PosEviLoc is
also best at every reported rank on KITTI360Pose, S3DISPose, and
NPM3DCoursePose. On PandaSetPose and nuScenesPose, it achieves the best R@1
while trailing the strongest R@2/R@3 result by only 0.02/0.05 and 0.04/0.03,
respectively. The large gain on S3DISPose suggests that QSEF remains effective
for cluttered indoor rooms with adaptive cell sizes, rather than relying on a
fixed outdoor grid. Its leading R@1 on PandaSetPose and nuScenesPose further
suggests consistent performance under dynamic traffic and in dense urban
layouts, while the NPM3DCoursePose gain extends to smaller cells. Together, these results outperform prior methods and demonstrate consistent performance across different scene types and cell constructions.

\subsection{Plug-and-Play Reranking}
\label{sec:plugin_reranking}

Table~\ref{tab:plugin_reranking} evaluates PosEviLoc as a plug-and-play
reranker for the Top-25 candidates produced by six retrieval methods, without
modifying or retraining their encoders. PosEviLoc improves 89 of the 90
reported R@1--R@3 results. The gains are especially large on S3DISPose, where
all methods improve by 30--37 percentage points. On NPM3DCoursePose, MNCL
increases from 0.10/0.16/0.19 to 0.32/0.46/0.56 at R@1/R@2/R@3. Consistent
improvements are also observed on the outdoor datasets, including R@1 gains of
0.28 and 0.27 for MambaPlace and CMMLoc on PandaSetPose.

These results demonstrate that position-conditioned object evidence
complements global cross-modal similarity. The base retriever efficiently
generates candidate submaps, while PosEviLoc improves their ordering by
verifying whether the objects jointly satisfy the query semantics and spatial
relations.

\begin{figure*}
    \centering
    \includegraphics[width=0.85\linewidth]{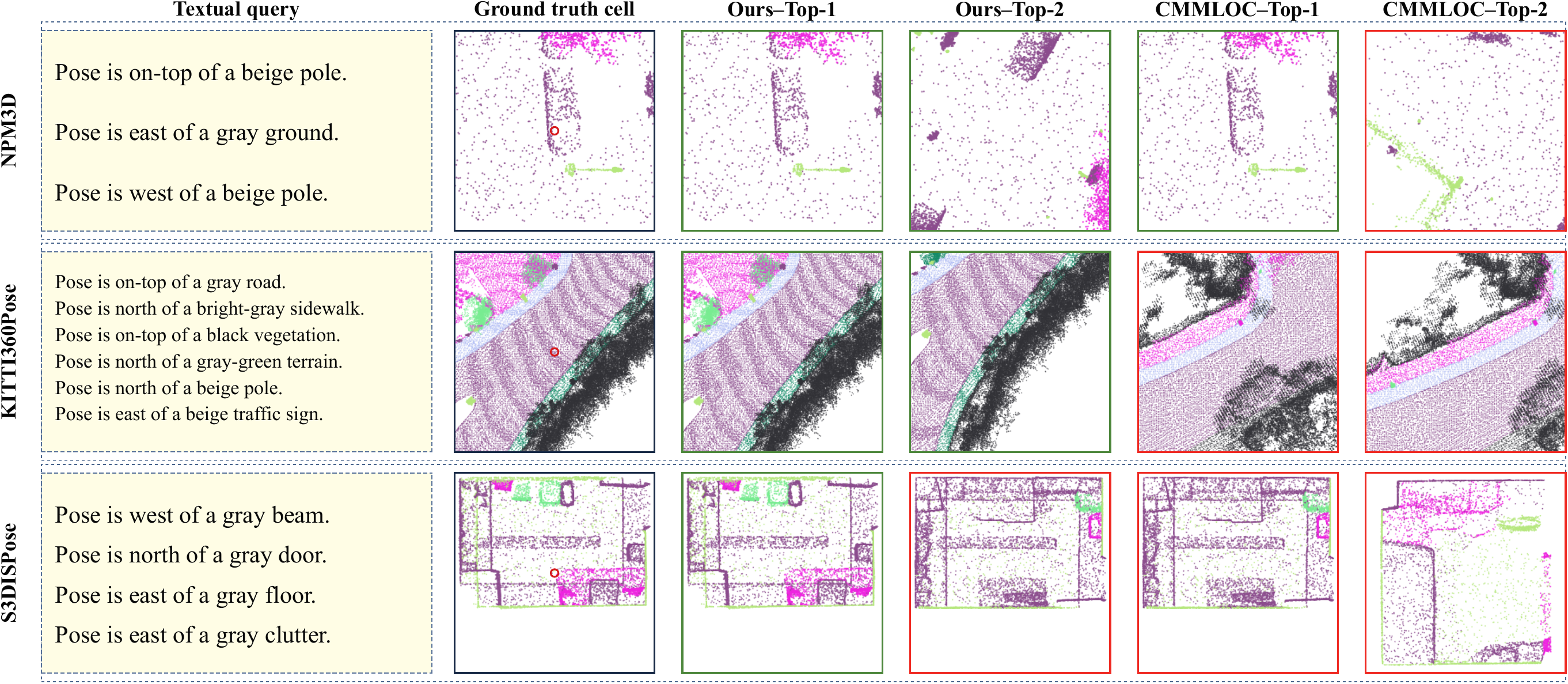}
    \caption{Qualitative retrieval results on 3 Datasets. From left to right: text query, ground-truth cell, Top-1 and Top-2 results from our method, and those from CMMLoc. Green and red borders indicate correct and incorrect retrievals, respectively.}
    \label{fig:retrieval_visualization}
\end{figure*}

\subsection{Ablation and Parameter Analysis}
Figure~\ref{fig:component_ablation} confirms that class, color, and direction
are complementary. Removing direction causes the largest average degradation,
particularly on PandaSetPose and NPM3DCoursePose, whereas removing class or
color also produces substantial losses across datasets. Disabling contextual
class refinement also reduces performance. More importantly, replacing QSEF
with directions computed only at the cell center sharply reduces S3DISPose
R@1 and lowers mean R@1--R@5. A single center-based direction assignment misses
valid query locations; evaluating direction over hypothetical positions is
therefore important for finding where multiple descriptions agree.

The MER results in Table~\ref{tab:mer_feature_ablation} show that the retained evidence coordinates are useful to the scorer. Removing $H$ degrades all datasets, reducing R@1 from 0.53 to 0.35 on KITTI360Pose, from 0.62 to 0.39 on nuScenesPose, and from 0.32 to 0.20 on NPM3DCoursePose. Since $H=I+A+D_{\max}$, removing it also eliminates the recoverable semantic term $I$ while retaining $A$ and $D_{\max}$, demonstrating the value of an aggregate semantic--spatial coordinate rather than an independent evidence source.

Removing the object-count prior $C$ causes drops, including 0.41 to 0.31 on S3DISPose and 0.50 to 0.41 on PandaSetPose, showing that scene-density context helps interpret the evidence. The soft term $A$ is selectively beneficial: its removal leaves three datasets unchanged but reduces KITTI360Pose from 0.53 to 0.36 and NPM3DCoursePose from 0.32 to 0.29, consistent with retaining partial class--color--direction matches under attribute errors.

Finally, removing $D_{\max}$ lowers S3DISPose from 0.41 to 0.31, whereas removing $D_{\mu}$ reduces S3DISPose, PandaSetPose, and nuScenesPose to 0.36, 0.45, and 0.57. These results show that peak exact feasibility and the average distribution of exact support capture different properties of QSEF.

\begin{figure}
    \centering
    \includegraphics[width=1\linewidth]{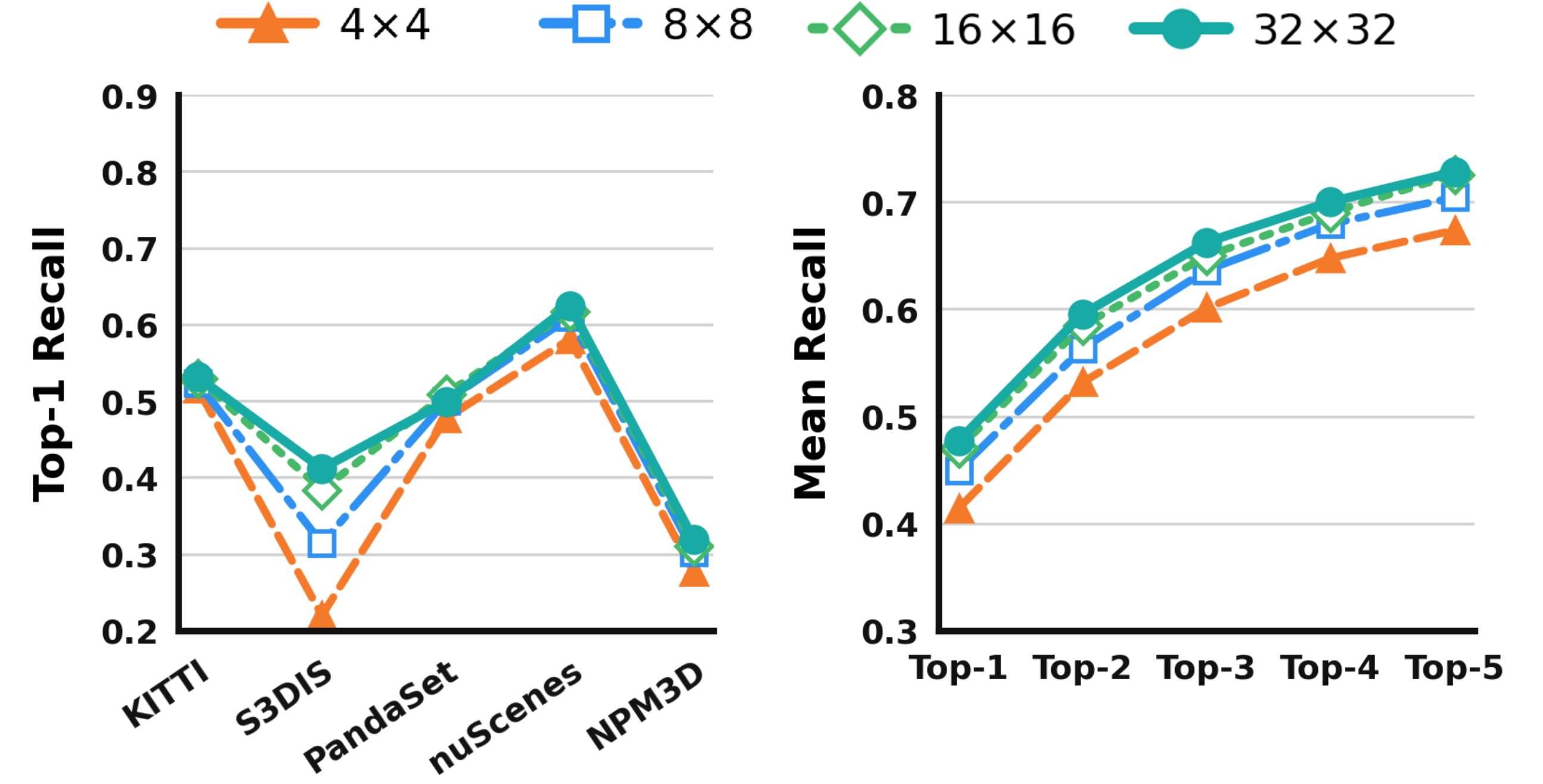}
    \caption{Effect of QSEF grid resolution. Left: Recall@1 by dataset. Right: mean Recall@$K$ across datasets for $K=1,\ldots,5$.}
    \label{fig:grid_ablation}
\end{figure}

We further analyze the impact of QSEF grid resolution. As shown in Fig.~\ref{fig:grid_ablation}, different resolutions are evaluated on five datasets. The results show that increasing the grid resolution generally improves retrieval performance. We therefore adopt 32×32 as the default setting.

\subsection{Visualization}
Figure~\ref{fig:retrieval_visualization} visualizes the Top-1 and Top-2
retrievals for three representative queries. PosEviLoc ranks the ground-truth
cell first in all examples, whereas CMMLoc places spatially inconsistent cells
first on KITTI360Pose and S3DISPose. 

\begin{figure}
    \centering
    \includegraphics[width=1\linewidth]{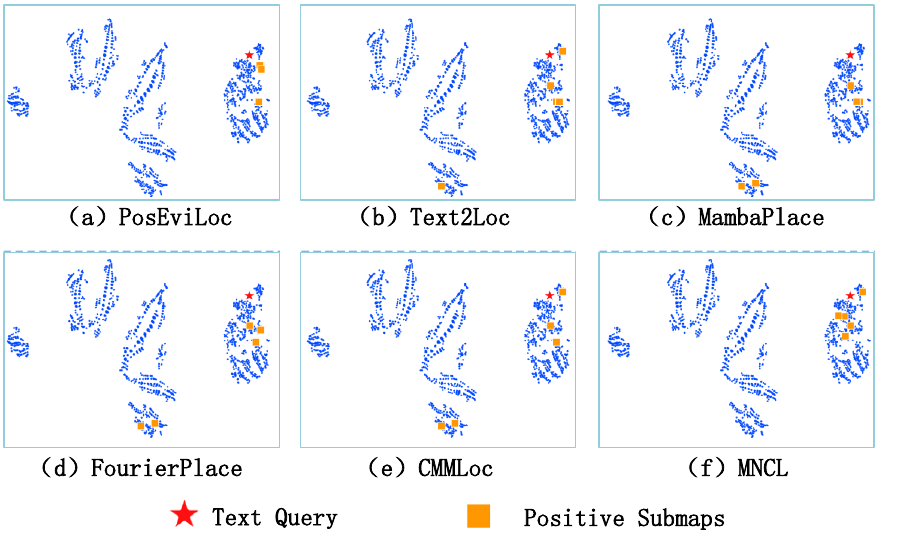}
    \caption{t-SNE visualization of the Top-5 retrieved submaps on KITTI360Pose.}
    \label{fig:tsne}
\end{figure}

The t-SNE visualization in Figure~\ref{fig:tsne} further suggests that the Top-5 submaps retrieved by PosEviLoc form a compact cluster around the text query and lie close to the ground-truth location. In contrast, the retrieval results of competing methods are more widely scattered, indicating weaker consistency between their retrieved submaps, the language query, and the target location.

\subsection{Computing Resource}
On KITTI360Pose, our method contains only 0.017M parameters, substantially fewer than MambaPlace (366.917M), CMMLoc (357.205M), and MNCL (359.029M), because it does not rely on heavyweight T5 backbone. Although the QSEF-based position hypotheses introduce additional computational overhead, our method requires only 7.09 ms per query, remaining clearly faster than MambaPlace (26.48 ms), CMMLoc (12.37 ms), and MNCL (14.46 ms). Therefore, our method provides clear advantages in both model size and inference speed.

\section{Conclusion}
This paper presented PosEviLoc, an evidence-based framework for
text-to-point-cloud localization. PosEviLoc treats direction as a
relation between an object and a hypothetical query position. QSEF evaluates
class, color, and direction support over candidate positions, thereby measuring
whether multiple descriptions agree at a shared location, while MER combines
complementary semantic and spatial evidence for retrieval scoring. Across five benchmarks, PosEviLoc improves Recall@1 over the strongest baseline by 0.10, 0.31, 0.40, 0.10, and 0.13, respectively, while improving 89 of 90 reranking results. Moreover, it requires substantially fewer parameters and achieves faster inference than previous methods.

\begingroup
\small
\bibliography{aaai2027}
\endgroup
\clearpage
\onecolumn
\raggedbottom
\appendix
\setcounter{secnumdepth}{2}
\setcounter{figure}{0}
\setcounter{table}{0}
\setcounter{equation}{0}
\renewcommand{\thefigure}{A\arabic{figure}}
\renewcommand{\thetable}{A\arabic{table}}
\renewcommand{\theequation}{A\arabic{equation}}
\renewcommand{\theHfigure}{appendix.\arabic{figure}}
\renewcommand{\theHtable}{appendix.\arabic{table}}
\renewcommand{\theHequation}{appendix.\arabic{equation}}
\section*{Appendix}
\addcontentsline{toc}{section}{Appendix}
\section{Independent Method Comparison}

\begin{figure}[H]
    \centering
    \includegraphics[width=0.9\linewidth,height=0.59\textheight,keepaspectratio]{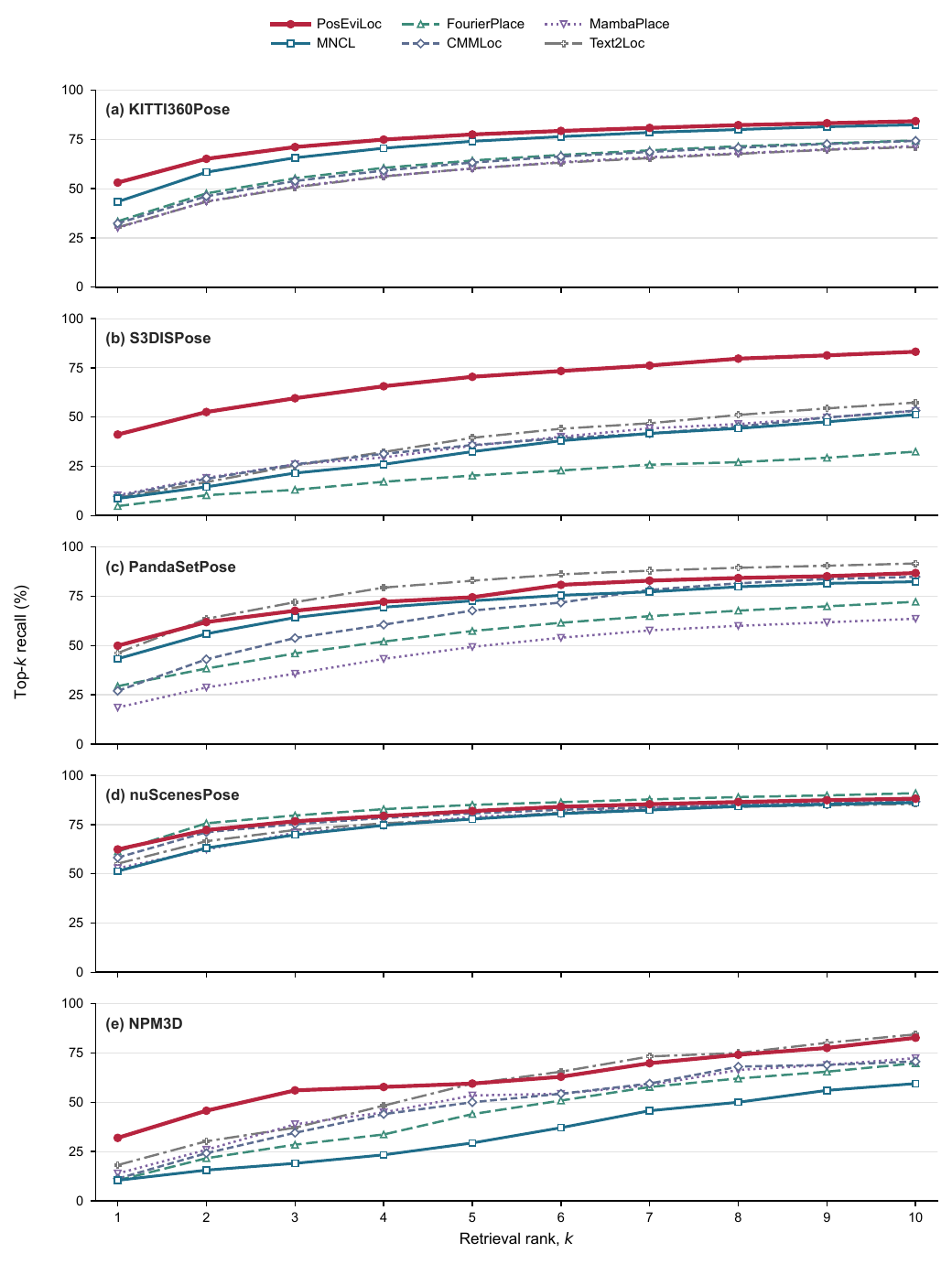}
     \caption{Top-\(k\) place-retrieval performance on five datasets. We compare PosEviLoc with five baseline methods on KITTI360Pose, S3DISPose, PandaSetPose, nuScenesPose, and NPM3D in terms of Top-1 through Top-10 recall. A query is considered successfully localized when its ground-truth submap cell appears among the top \(k\) retrieved candidates. PosEviLoc achieves superior or competitive recall across most datasets and retrieval depths, with particularly clear improvements on KITTI360Pose, S3DISPose, and NPM3D.}
    \label{Fig:Appendix_Figure1_Performance_Compare}
\end{figure}

Figure \ref{Fig:Appendix_Figure1_Performance_Compare} presents the complete Top-1 to Top-10 retrieval curves of PosEviLoc and five competing methods, namely Text2Loc, FourierPlace, MambaPlace, CMMLoc, and MNCL, across five datasets. In contrast to the individual Top-\(k\) metrics reported in the main paper, these curves characterize how recall evolves as the retrieval depth increases. All results are evaluated at the submap-cell level: a query is counted as correct if its ground-truth submap cell is included among the top \(k\) retrieved candidates.
On KITTI360Pose, PosEviLoc consistently outperforms the baselines across all retrieval depths, with a particularly clear margin at small values of \(k\). This indicates that PosEviLoc is more effective at placing the correct location near the top of the ranked candidate list. On S3DISPose, PosEviLoc delivers a substantial and consistent improvement over all competing methods. It also achieves the best performance at most retrieval depths on NPM3D, with pronounced gains from Top-1 to Top-5. 

On PandaSetPose, PosEviLoc outperforms most baselines, although Text2Loc obtains higher recall at larger retrieval depths. The performance differences are smaller on nuScenesPose, where PosEviLoc nevertheless remains competitive. Overall, the results demonstrate that the advantage of PosEviLoc is not confined to a particular Top-\(k\) threshold; instead, the method maintains robust ranking performance across diverse environments and retrieval depths.

\clearpage

\section{Comparison as Plug-and-play Module}

\begin{figure}[H]
    \centering
    \includegraphics[width=1\linewidth,height=0.17\textheight,keepaspectratio]{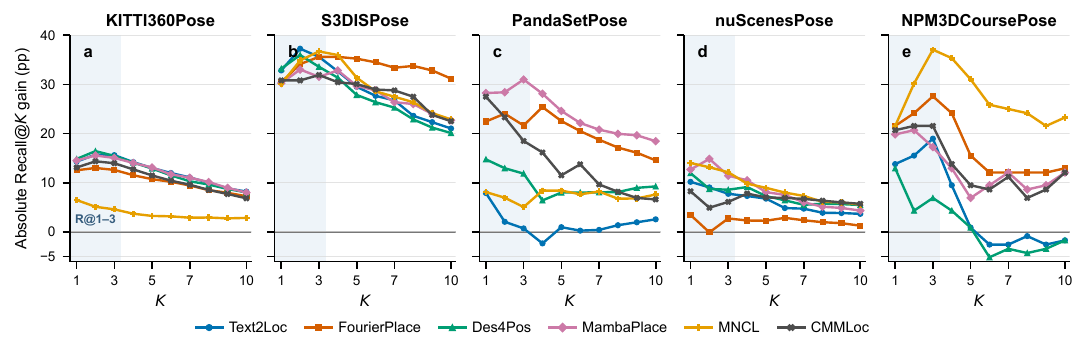}
     \caption{Absolute Recall@$K$ improvement produced by PosEviLoc. Each panel shows the change in recall after applying PosEviLoc to the
Top-25 candidates retrieved by six coarse localization methods on one
dataset. We report absolute improvement as
$\Delta R@K=(R@K_{\mathrm{reranked}}-R@K_{\mathrm{original}})\times100$
percentage points (pp). The shaded region denotes $K=1$--$3$ and
corresponds to the operating points reported in the paper. PosEviLoc changes
only the ordering of the original Top-25 candidates; their membership
remains fixed.}
    \label{Fig:rankwise_reranking}
\end{figure}

\begin{figure}[H]
    \centering
    \includegraphics[width=1\linewidth,height=0.22\textheight,keepaspectratio]{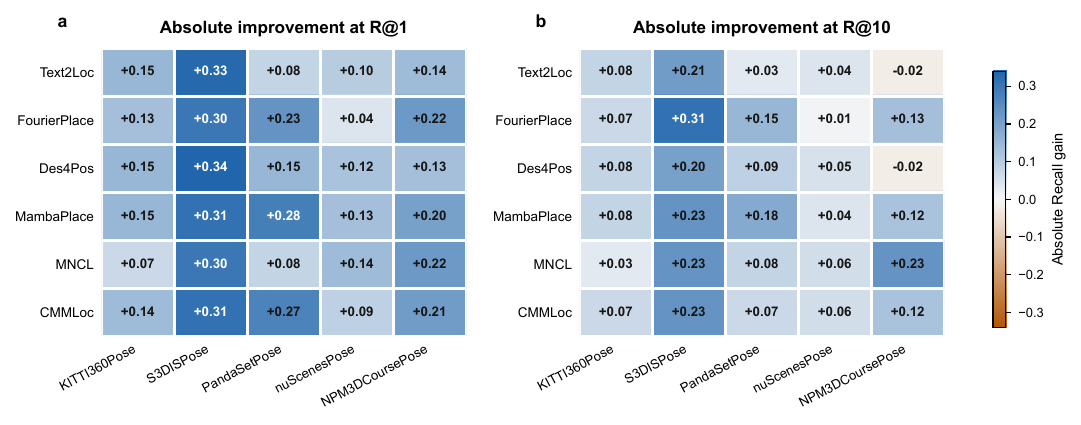}
     \caption{Absolute recall improvement at the leading and broader
retrieval cutoffs.
Left, Absolute $R@1$ improvement from PosEviLoc for each coarse
method--dataset pair.
Right, Corresponding absolute $R@10$ improvement.
Each cell reports
$\Delta R@K=R@K_{\mathrm{reranked}}-R@K_{\mathrm{original}}$
as a recall fraction (0.01 corresponds to one percentage point), computed from full-precision recall values before
display rounding. Blue cells indicate gains and orange cells indicate
decreases. The two panels use the same color scale, and the Top-25
candidate membership is fixed throughout reranking.}
    \label{Fig:rankwise_reranking_heatmap}
\end{figure}

As shown in Figures~\ref{Fig:rankwise_reranking} and~\ref{Fig:rankwise_reranking_heatmap}, PosEviLoc improves $R@1$ for all 30 method--dataset combinations. Based on
the recall fractions displayed in the heatmap, the mean absolute improvement is 0.18,
with individual gains ranging from 0.04 to 0.34. The largest and most
consistent improvements occur on S3DISPose, where all six coarse retrievers
gain between 0.30 and 0.34 at $R@1$. Substantial gains are also observed on
PandaSetPose and NPM3DCoursePose, although their magnitude varies more across
the coarse retrieval methods. This consistency across six substantially
different retrievers indicates that the reranking benefit is not restricted
to one retrieval architecture.

The improvement remains positive for 28 of the 30 method--dataset
combinations at $R@10$, with a mean change of 0.11. The lower average gain
at $R@10$ than at $R@1$ indicates that PosEviLoc primarily changes the
ordering near the head of the retrieved list. S3DISPose again shows the
strongest overall response, with $R@10$ gains between 0.20 and 0.31. On
NPM3DCoursePose, MNCL obtains a gain of 0.23, whereas Text2Loc and Des4Pos
each decrease by 0.02. These two decreases represent the only negative cells
in the figure.

The isolated $R@10$ decreases do not indicate a reduction in Top-25
candidate coverage. PosEviLoc operates on a fixed Top-25 candidate set, so a
negative $\Delta R@10$ means that, for a small subset of queries, the correct
candidate is moved from the first ten positions to a lower position within
the same candidate set. Conversely, the broad positive pattern at $R@1$ and
$R@10$ shows that PosEviLoc usually moves correct candidates toward the
leading ranks. The results therefore support the intended role of PosEviLoc
as a candidate-preserving reranker rather than an independent retrieval
stage.

\clearpage

\section{Visualization Results for Right Cases}
\begin{figure}[H]
    \centering
    \includegraphics[width=0.8\linewidth,height=0.73\textheight,keepaspectratio]{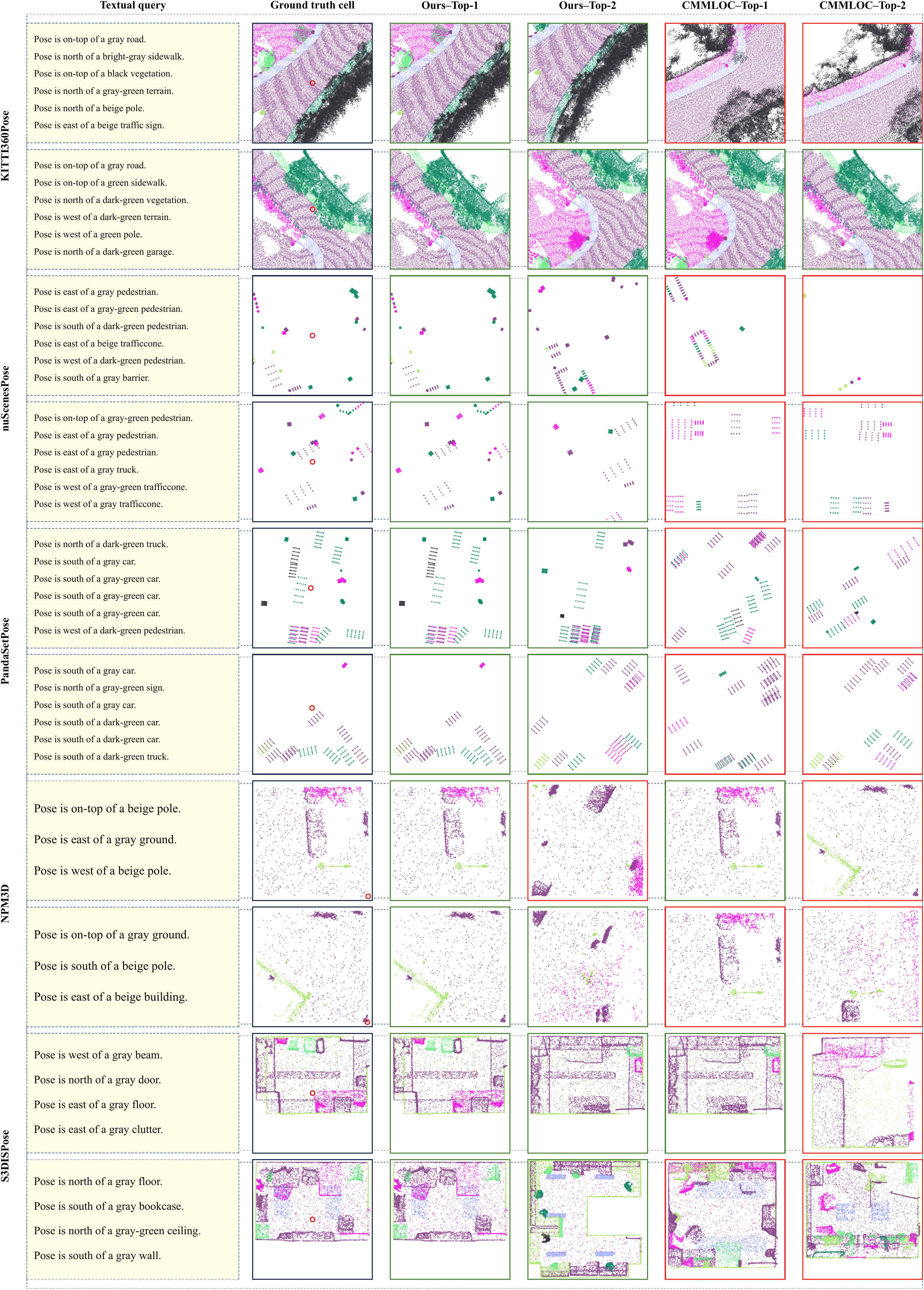}
     \caption{Two query examples are shown for each dataset. From left to right, the columns present the complete textual query, the ground-truth cell, the Top-1 and Top-2 cells retrieved by our method, and the corresponding Top-1 and Top-2 results of CMMLOC. Green and red borders denote correct and incorrect retrievals, respectively.}
    \label{Fig:Reterival_Results_Visulization_right}
\end{figure}

This Figure~\ref{Fig:Reterival_Results_Visulization_right} presents qualitative retrieval results of our method across multiple datasets. Despite substantial differences in scene structure, point-cloud density, and semantic distribution, our method accurately retrieves the target cells by leveraging the semantic and spatial relations expressed in textual descriptions. In contrast, CMMLOC fails to retrieve the correct locations in most cases. These results intuitively demonstrate the superior accuracy and robustness of our method across multiple datasets.

\clearpage

\section{Visualization Results for Wrong Cases}
\begin{figure}[H]
    \centering
    \includegraphics[width=0.8\linewidth,height=0.59\textheight,keepaspectratio]{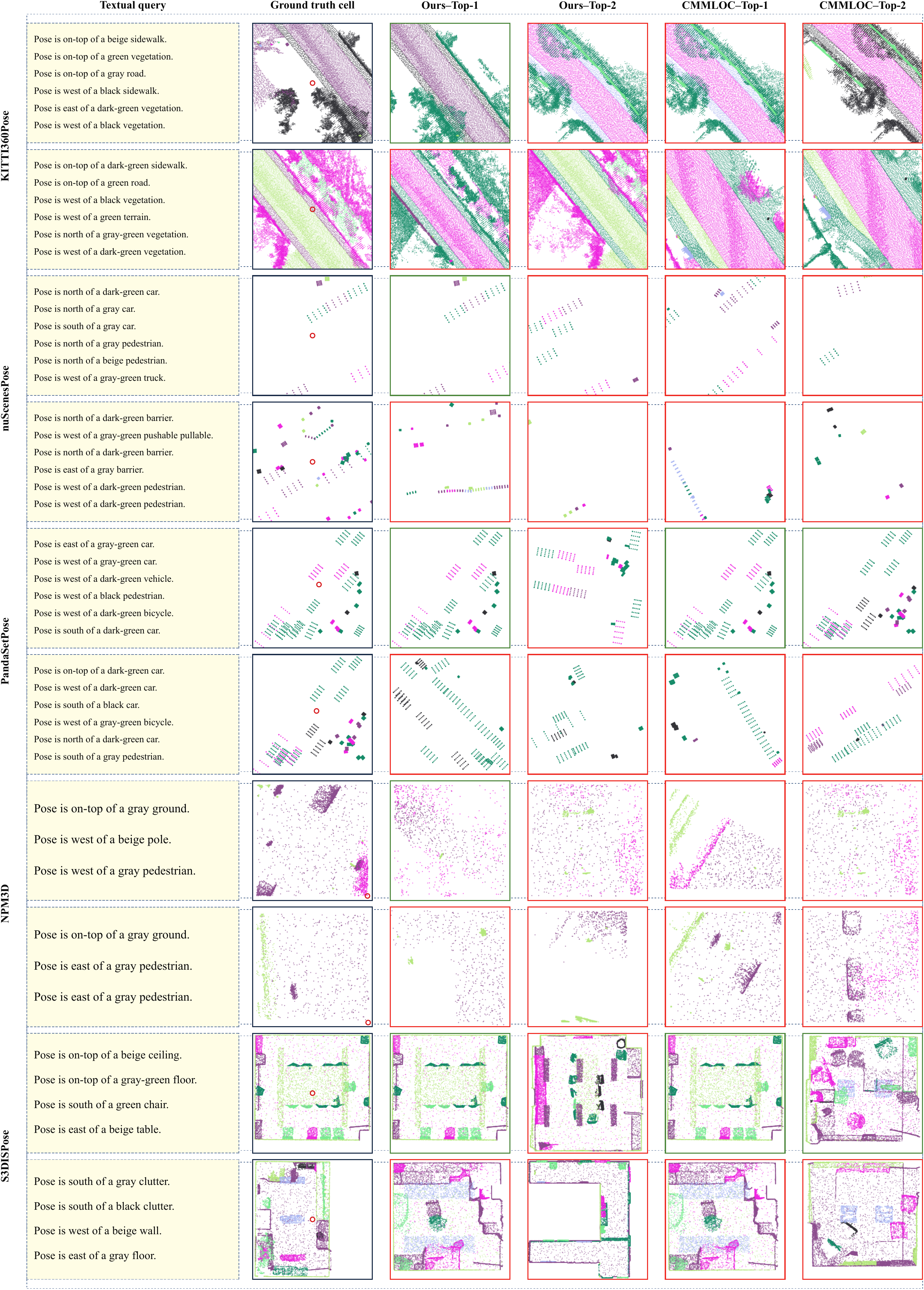}
     \caption{Two query examples are shown for each dataset. From left to right, the columns present the complete textual query, the ground-truth cell, the Top-1 and Top-2 cells retrieved by our method, and the corresponding Top-1 and Top-2 results of CMMLOC. Green and red borders denote correct and incorrect retrievals, respectively.}
    \label{Fig:Reterival_Results_Visulization_wrong}
\end{figure}

As shown in Figure~\ref{Fig:Reterival_Results_Visulization_wrong}, these examples are deliberately selected to analyze the retrieval behavior of our model under challenging conditions. In the first type of case, the ground-truth cell is ranked first, while an incorrect cell still receives a high similarity score and appears in the Top-2 results. This suggests that the model captures the dominant scene cues but does not always fully distinguish between candidates with similar semantic compositions. In the second type, neither of the top two candidates matches the ground-truth location, indicating stronger ambiguity between the query and the scene. Several factors may contribute to these errors: different locations may contain repetitive or highly similar road, building, or indoor structures; the landmark categories and color attributes mentioned in the query may provide insufficient discriminative information; spatial relations defined with respect to local reference objects can be confused across similar layouts; and cell-based spatial discretization may reduce matching consistency when the ground-truth position lies close to a cell boundary. Moreover, when a query contains multiple spatial constraints, an unstable or noisy relation may weaken the joint representation, causing the model to emphasize shared semantic elements while overlooking fine-grained geometric cues that distinguish candidate locations. These observations indicate that further improvements are possible in repetitive-scene discrimination, fine-grained spatial reasoning, and joint modeling of multiple spatial relations. The selected examples illustrate representative failure modes and are not intended to reflect the overall quantitative performance of the model
.

\clearpage

\section{Visualization for Top-5 Submaps}
\begin{figure}[H]
    \centering
    \includegraphics[width=0.8\linewidth,height=0.67\textheight,keepaspectratio]{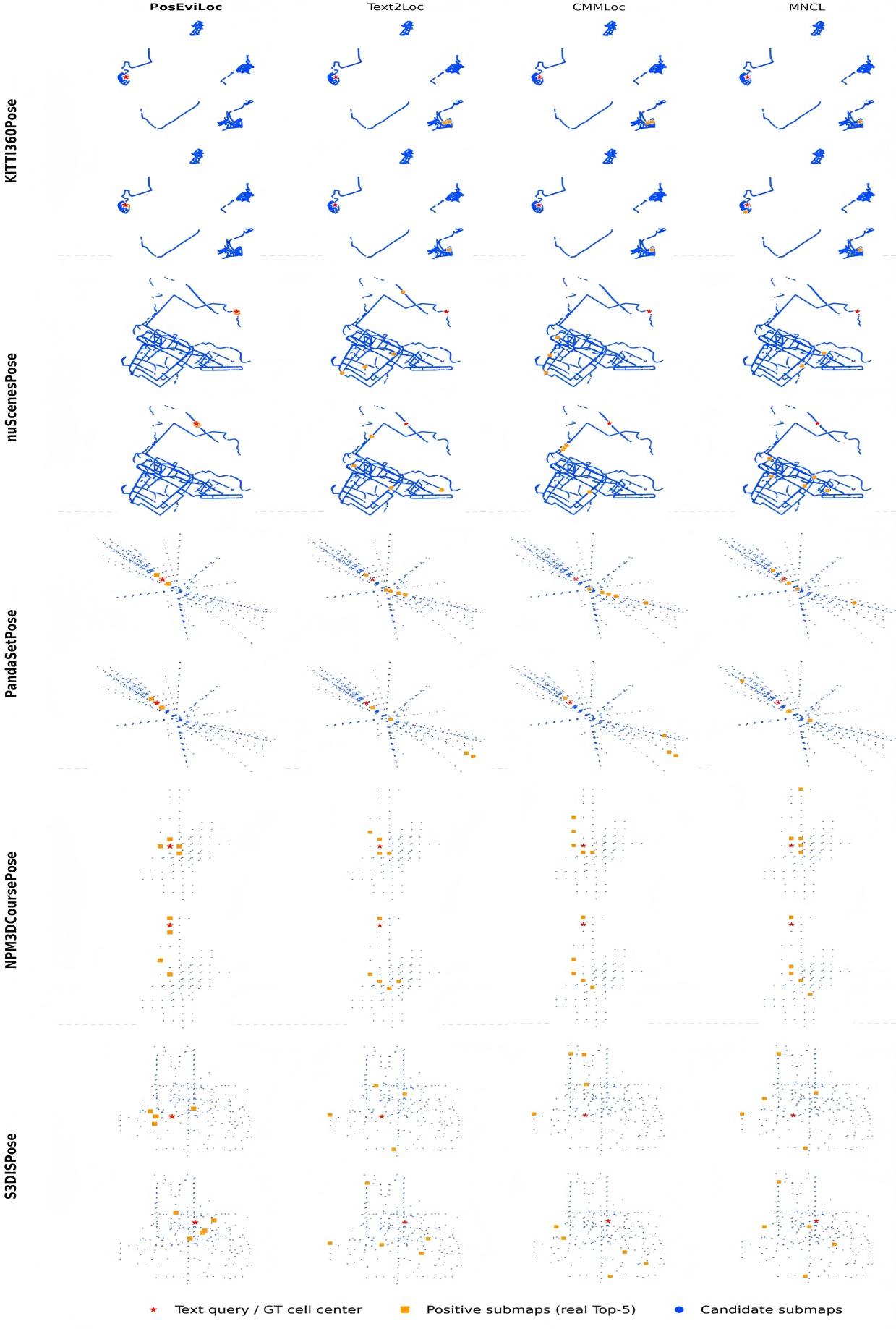}
     \caption{Real Top-5 retrieval visualization of PosEviLoc and six baselines on five datasets. Each dataset contains two fixed query scenes, and each row compares the same query across seven methods. Blue points denote the real world-coordinate centers of test submaps, the red star denotes the GT cell center corresponding to the text query, and orange squares denote the real Top-5 retrieved submaps of each method.}
    \label{Fig:Top-5-Distributation}
\end{figure}

To further examine the spatial behavior of different methods in coarse submap retrieval, we visualize the real Top-5 retrieval results for fixed query scenes across five datasets. Unlike schematic t-SNE visualizations, this figure directly uses the real world-coordinate centers of submaps from each dataset, so the spatial distance between the orange retrieved submaps and the red query location directly reflects localization error. As shown in Figure~\ref{Fig:Top-5-Distributation} for each dataset, we select two real queries for visualization, where PosEviLoc achieves smaller average spatial distances than the six baselines. The results show that PosEviLoc more reliably retrieves submaps near the target query location, while the baseline methods often include spatially distant candidates in their Top-5 results, even when some retrieved submaps are close to the target. This demonstrates that PosEviLoc provides more stable spatial evidence modeling and stronger coarse localization robustness across datasets.

\clearpage

\section{Visualization of the Five-Dimensional Features}
\begin{figure}[H]
    \centering
    \includegraphics[width=0.8\linewidth,height=0.5\textheight,keepaspectratio]{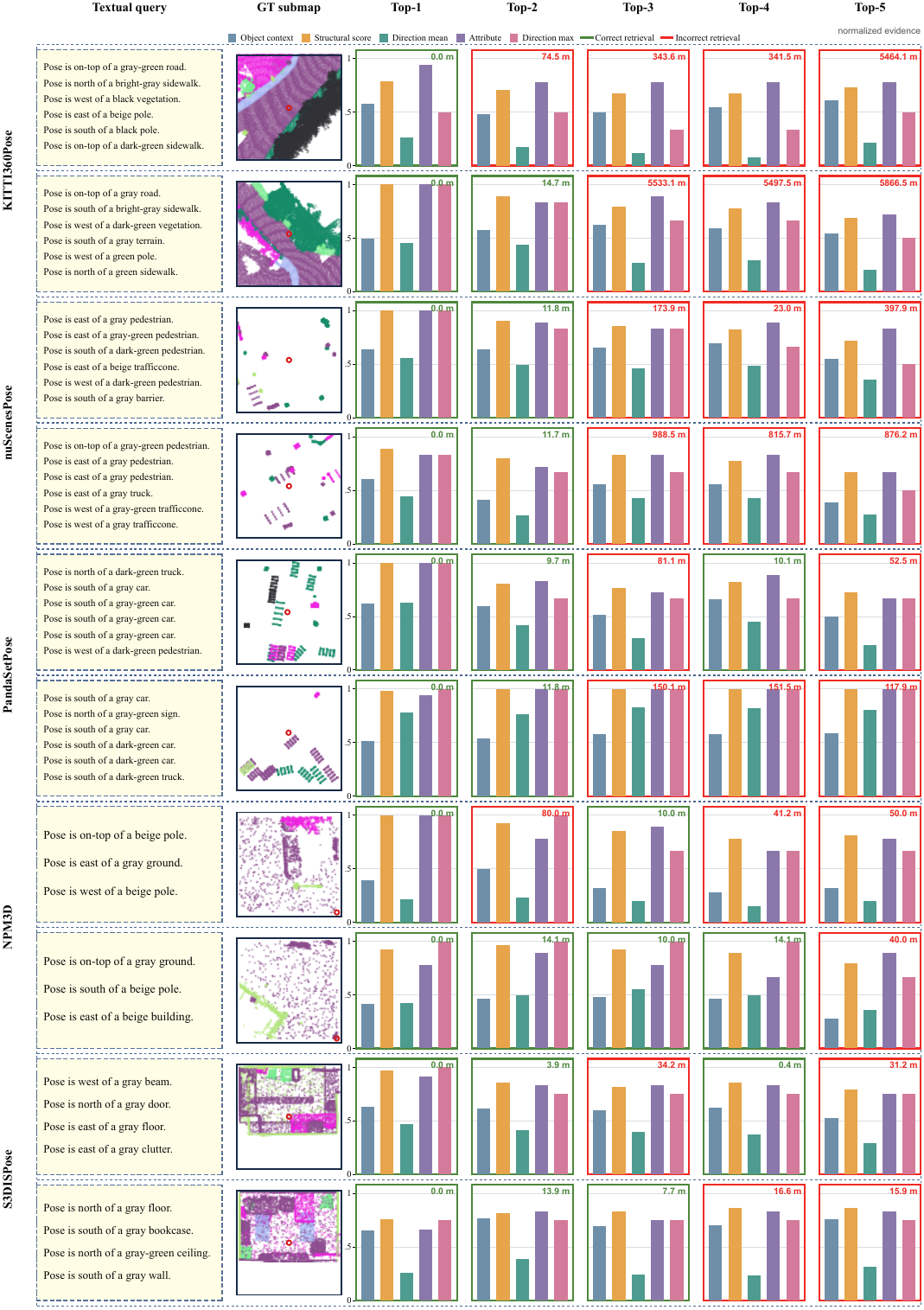}
     \caption{Visualization of Top-5 retrieval results and five-dimensional features across five datasets. Two textual queries are presented for each dataset. From left to right, the columns show the complete textual query, the ground-truth submap, and the Top-1 to Top-5 candidates retrieved by our method. For each candidate, the bar chart visualizes five retrieval features: object context, structural score, direction mean, attribute matching, and direction maximum. For visualization, the structural score is divided by three, while the other features retain their original model scales. Green and red borders indicate correct and incorrect retrievals, respectively. The value in the upper-right corner reports the distance between the center of the candidate submap and the ground-truth location.}
    \label{Fig:five_feature}
\end{figure}

As shown in Figure~\ref{Fig:five_feature}, this figure further illustrates how the five-dimensional features contribute to cross-dataset retrieval. Correct candidates generally exhibit strong and consistent responses across structural matching, attribute matching, and directional consistency. This observation suggests that our method does not rely solely on the presence of semantically matched objects but jointly considers their attributes and spatial configurations. In contrast, some incorrect candidates achieve relatively high object-context or attribute scores but show weaker mean directional consistency. Such candidates contain semantically similar objects but fail to satisfy multiple spatial constraints expressed by the query. Therefore, individual semantic or attribute evidence is insufficient to reliably distinguish visually similar scenes, whereas the complementary five-dimensional representation supports more accurate candidate ranking.

The feature distributions also vary across datasets. Sparse driving scenes depend more strongly on spatial relations among a limited number of landmarks, whereas dense outdoor and indoor environments provide richer object and attribute evidence. Despite substantial differences in scene structure, point-cloud density, and semantic composition, correct candidates generally exhibit more coherent combinations of multidimensional evidence. This result demonstrates the cross-dataset adaptability of the proposed features. Nevertheless, some incorrect candidates present feature profiles similar to those of correct candidates, indicating that repetitive scene structures, common semantic objects, and ambiguities near submap boundaries can still lead to retrieval confusion.

\clearpage

\section{Dataset Details}
\begin{figure}[H]
    \centering
    \includegraphics[width=1\linewidth,height=0.86\textheight,keepaspectratio]{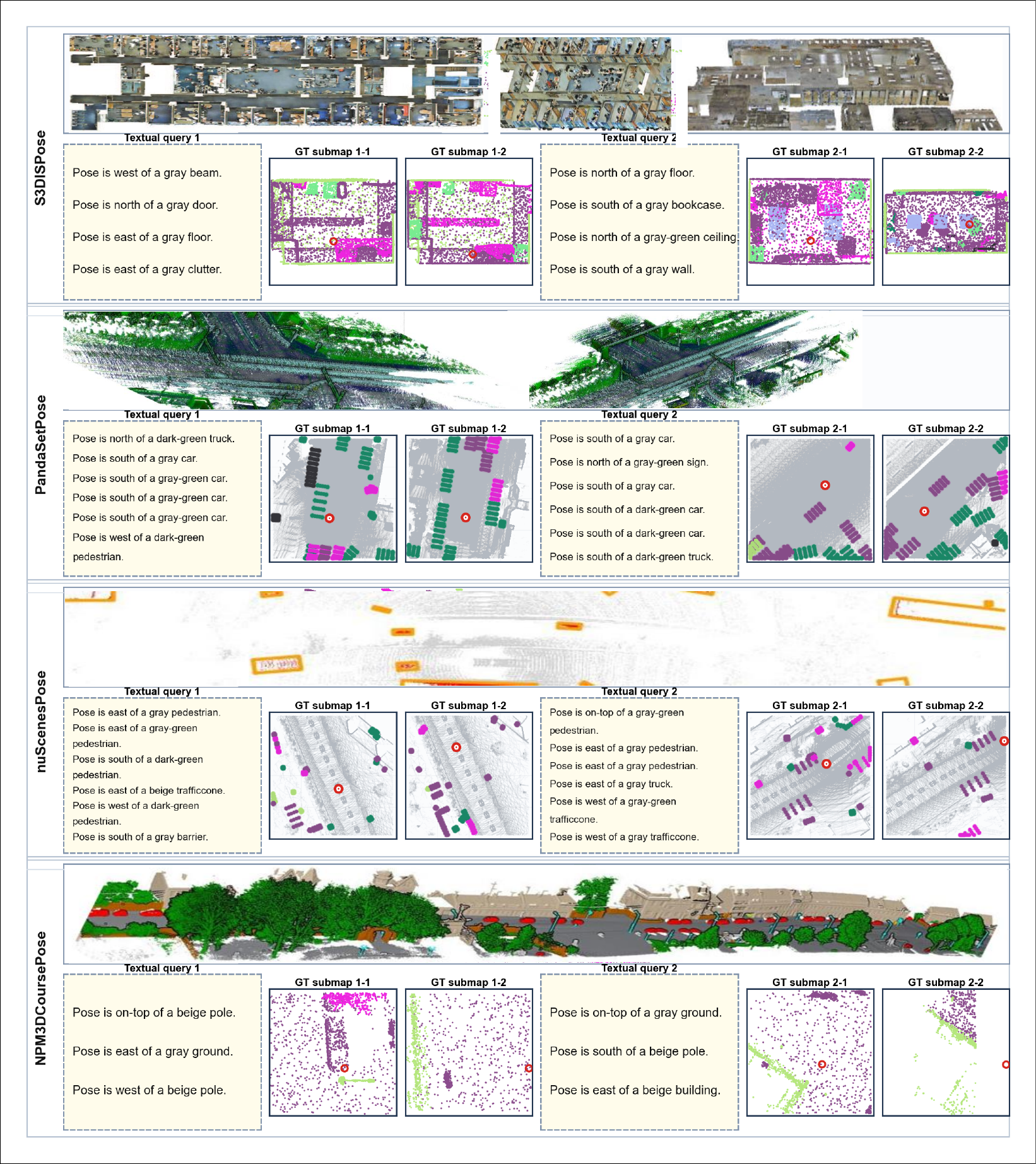}
     \caption{Overview of the four constructed pose-localization datasets. From top to bottom, we show S3DISPose, PandaSetPose, nuScenesPose, and NPM3DCoursePose. For each dataset, the upper row presents a large-scale point-cloud overview, while the lower row contains two textual queries, each paired with the annotated ground-truth submap and an additional valid ground-truth submap whose center lies within 15m of the target pose.}
    \label{Fig:Datasets}
\end{figure}

To evaluate text-to-pose localization beyond KITTI360Pose, we construct four additional benchmarks from S3DIS, PandaSet, nuScenes, and Paris-Lille-3D. Although these source datasets differ substantially in sensing platforms, spatial scales, and annotation formats, we convert them into a unified KITTI360Pose-compatible representation. The construction pipeline comprises coordinate alignment, semantic-instance extraction, submap generation, pose sampling, spatial-language generation, and leakage-free data splitting. Each resulting sample contains a query pose, a natural-language description of nearby landmarks, and its associated ground-truth submap. In Figure~\ref{Fig:Datasets}, we have provided a detailed presentation of the four newly established datasets we have created.

\subsection{S3DISPose.} S3DISPose is constructed from the aligned version of S3DIS and represents indoor localization in offices, conference rooms, corridors, and other building environments. Each complete room is treated as an independent scene and converted into one adaptive submap whose spatial extent covers the entire room. Object instances are obtained from the original instance-level annotations and retain their semantic categories and RGB colors. We sample up to eight query poses per room and describe each pose using four nearby semantic landmarks and their relative directions. To prevent room-level overlap, complete rooms are deterministically divided into 136 training, 68 validation, and 68 test rooms. The resulting benchmark contains 272 submaps, 2,170 poses, and 9,832 semantic objects.

\subsection{PandaSetPose.} PandaSetPose extends the task to large-scale urban-driving environments captured by a roof-mounted LiDAR system. LiDAR measurements and annotated objects are represented in a common world coordinate frame using the provided sensor poses. We construct overlapping \(30\,\mathrm{m}\times30\,\mathrm{m}\) submaps at a spatial interval of 10m and aggregate semantic instances observed within each local region. Query poses follow the vehicle trajectory, and each query is described using six nearby landmarks, including vehicles, pedestrians, traffic cones, signs, barriers, and other traffic-related objects. The 100 complete driving sequences are divided by sequence into 50 training, 25 validation, and 25 test sequences, thereby avoiding frame-level leakage between splits. PandaSetPose contains 779 submaps, 8,000 query poses, and 11,789 semantic objects.

\subsection{nuScenesPose.} nuScenesPose is constructed from the 850 scenes of the nuScenes v1.0-trainval release. Raw LiDAR points are first transformed from the LiDAR sensor frame to the ego-vehicle frame and subsequently to the global world frame using the calibrated-sensor and ego-pose transformations. Overlapping 30m submaps are generated at 10m intervals along the vehicle trajectories. Six nearby annotated instances are selected for each query pose to form spatial descriptions covering pedestrians, vehicles, barriers, traffic cones, bicycles, and other urban objects. Splitting is performed at the complete-scene level: 425 scenes are used for training, 212 for validation, and 213 for testing. The test set includes all 150 official nuScenes validation scenes and 63 deterministically selected training scenes. The resulting benchmark comprises 6,944 submaps, 30,151 query poses, and 53,044 semantic objects.

\subsection{NPM3DCoursePose.} NPM3DCoursePose is derived from the Paris-Lille-3D/NPM3D mobile-laser-scanning benchmark and evaluates localization in dense outdoor street environments. We retain six landmark categories—ground, building, pole, pedestrian, car, and vegetation—and partition the source scans into overlapping 20m submaps with a spacing of 10m. Each retained cell contains at least three usable semantic landmarks, and four query poses are sampled per cell. Each query is described using three nearby landmarks and their spatial relations. Because the source point clouds do not provide RGB values, deterministic semantic pseudo-colors are assigned only for visualization. Geographical separation is maintained by using Lille1 for training, Lille2 for validation, and Paris for testing. The constructed dataset contains 91 submaps, 364 query poses, and 417 semantic objects.

\clearpage

\clearpage

\section{Five-seed stability analysis}
\begin{figure}[H]
    \centering
    \includegraphics[width=1\linewidth,height=0.2\textheight,keepaspectratio]{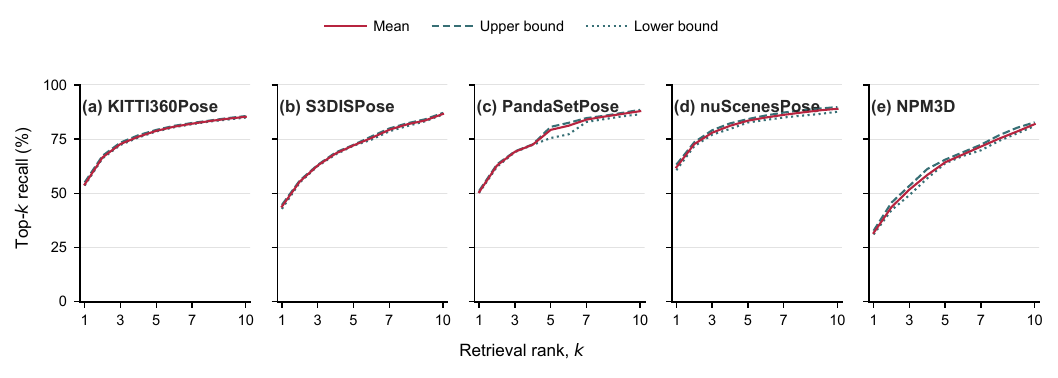}
     \caption{Five-seed stability analysis of PosEviLoc across five datasets. Top-1 through Top-10 recall is reported on KITTI360Pose, S3DISPose, PandaSetPose, nuScenesPose, and NPM3D. The solid line denotes the mean performance over five random seeds, while the upper and lower boundaries indicate the maximum and minimum recall observed at each retrieval depth, respectively.}
    \label{Fig:5_Seeds}
\end{figure}

Figure~\ref{Fig:5_Seeds} examines the sensitivity of PosEviLoc to training randomness. Across all five datasets, the mean recall increases consistently with retrieval depth, while the upper and lower boundaries remain relatively close at most values of \(k\). This indicates that the reported performance does not depend on a particular random initialization.
KITTI360Pose exhibits the highest stability among the five datasets. Its Top-1, Top-5, and Top-10 recalls are \(54.07\pm0.61\%\), \(78.79\pm0.27\%\), and \(85.29\pm0.25\%\), respectively. The upper and lower boundaries nearly overlap across the complete Top-1 to Top-10 range, indicating that model initialization and negative sampling have only a limited effect on the final rankings for this large-scale test set.
PosEviLoc is also stable on S3DISPose, achieving Top-1 and Top-10 recalls of \(43.87\pm0.74\%\) and \(86.61\pm0.31\%\), respectively. Although this dataset contains only 542 test queries, the retrieval curves remain consistent across the five runs. This suggests that the model reliably learns the object attributes and spatial relationships present in indoor environments.
On PandaSetPose, the variation is small from Top-1 to Top-4 but becomes more pronounced at Top-5 and Top-6. The corresponding standard deviations are 2.13 and 2.24 percentage points, primarily because seed 3 performs relatively poorly at these two retrieval depths. The difference narrows again as \(k\) increases, resulting in a Top-10 recall of \(87.89\pm0.82\%\). The observed variation therefore mainly affects the local ordering of the correct candidate rather than its overall inclusion in the retrieved set.
nuScenesPose achieves the highest mean Top-1 recall, \(61.98\pm1.21\%\), among the five datasets. Its upper and lower boundaries are wider than those observed on KITTI360Pose, suggesting that the learned scorer is moderately more sensitive to training randomness on nuScenesPose. Nevertheless, the Top-10 recall remains consistently high at \(88.97\pm0.87\%\). Thus, the variation mainly influences the precise placement of the correct location near the top of the ranked list rather than whether it is retrieved within a broader candidate set.
On NPM3D, the Top-1 and Top-10 recalls are \(31.72\pm0.72\%\) and \(81.90\pm0.61\%\), respectively. This dataset contains only 116 test queries, meaning that a single query changes recall by approximately 0.86 percentage points. Its comparatively visible boundary width should therefore be interpreted in the context of the limited test-set size rather than being attributed entirely to unstable model training.
Overall, PosEviLoc demonstrates good reproducibility across random seeds. KITTI360Pose and S3DISPose produce the most stable results, PandaSetPose exhibits localized variation at intermediate retrieval depths, nuScenesPose shows moderate ranking sensitivity, and the variation on NPM3D is amplified by its small test set. These results confirm that the principal performance trends of PosEviLoc remain consistent under different model initializations and training orders.

\end{document}